\documentclass{article}
\usepackage{subcaption, booktabs, siunitx, adjustbox, bm}
\usepackage{mathrsfs}
\usepackage{algorithm}
\usepackage{algpseudocode}
\usepackage{amsmath, amssymb, mathtools, nicefrac}
\usepackage{cite}

\usepackage{lipsum}
\usepackage{amsfonts}
\usepackage{graphicx}
\usepackage{epstopdf}
\ifpdf
  \DeclareGraphicsExtensions{.eps,.pdf,.png,.jpg}
\else
  \DeclareGraphicsExtensions{.eps}
\fi
\newcolumntype{C}{>{\centering\arraybackslash}m{2cm}}
\newcolumntype{N}{>{\centering\arraybackslash}m{2.5cm}}

\newcommand{\Px}{{P_{\mathbf{x}}}}
\newcommand{\Pg}{{P_{\mathbf{g}}}}
\newcommand{\Pz}{{P_{\mathbf{z}}}}

\newcommand\subscr[2]{#1_{\textup{#2}}}

\newcommand{\Vvg}{\subscr{V}{VG}}
\newcommand{\E}{\mathbb{E}}
\newcommand{\A}{\textbf{A}}
\newcommand{\x}{\textbf{x}}
\newcommand{\z}{\textbf{z}}
\newcommand{\B}{\textbf{B}}
\newcommand{\jfd}[1]{\textsc{JD}_{#1}}

\newcommand{\La}{\mathcal{L}_\alpha}
\newcommand{\Lk}{\mathcal{L}_k}
\newcommand{\btheta}{\bm{\theta}}
\newcommand{\kl}{\textsc{KL}}

\newcommand{\lagan}{\La\mathrm{-GAN}}
\newcommand{\boldlagan}{\bm{\lagan}}
\newcommand{\agan}{\alpha\mathrm{-GAN}}
\newcommand{\boldagan}{\bm{\agan}}
\newcommand{\adggan}{(\alpha_D, \alpha_G)\mathrm{-GAN}}
\newcommand{\boldadggan}{\bm{\adggan}}
\newcommand{\intx}[1]{\int_{\mathcal{X}} #1 \,d\mu}
\newcommand{\intr}[1]{\int_{\mathcal{R}} #1 \,d\mu}
\newcommand{\includeplot}[1]{\includegraphics[width=\textwidth, trim={1.17cm 0.5cm 1.5cm 1.4cm}, clip]{#1}}
\newcommand{\includeimages}[1]{\includegraphics[width=\textwidth, trim = {2cm 1.2cm 1.7cm 1.5cm}, clip]{#1}}

\newtheorem{theorem}{Theorem}
\newtheorem{definition}{Definition}
\newtheorem{lemma}{Lemma}
\newtheorem{remark}{Remark}
\newtheorem{proposition}{Proposition}
\newcommand{\qed}{\hfill $\blacksquare$}

\title{\bf A Unifying Generator Loss Function \\for Generative Adversarial Networks}

\author{Justin Veiner\thanks{Department of Mathematics and Statistics, Queen's University, Kingston, Ontario, Canada
  (justin.veiner@queensu.ca).}
\and Fady Alajaji\thanks{Department of Mathematics and Statistics, Queen's University, Kingston, Ontario, Canada
(fa@queensu.ca).}
\and Bahman Gharesifard\thanks{Department of Electrical and Computer Engineering, University of California, Los Angeles, CA, USA
(gharesifard@ucla.edu).}
}

\usepackage{amsopn}

\allowdisplaybreaks

\begin{document}
\date{}
\maketitle

\begin{abstract}
    A unifying $\alpha$-parametrized generator loss function is introduced for a dual-objective generative adversarial network (GAN), which uses a canonical (or classical) discriminator loss function such as the one in the original GAN (VanillaGAN) system. The generator loss function is based on a symmetric class probability estimation type function, $\La$, and the resulting GAN system is termed $\La$-GAN. Under an optimal discriminator, it is shown that the generator's optimization problem consists of minimizing a Jensen-$f_\alpha$-divergence, a natural generalization of the Jensen-Shannon divergence, where $f_\alpha$ is a convex function expressed in terms of the loss function $\mathcal{L}_\alpha$. It is also demonstrated that this $\lagan$ problem recovers as special cases a number of GAN problems in the literature, including VanillaGAN, Least Squares GAN (LSGAN), Least $k$th order GAN (L$k$GAN) and the recently introduced $(\alpha_D,\alpha_G)$-GAN with $\alpha_D=1$. Finally, experimental results are conducted on three datasets, MNIST, CIFAR-10, and Stacked MNIST to illustrate the performance of various examples of the $\lagan$ system.
    \end{abstract}


\section{Introduction}
Generative adversarial networks (GANs), first introduced by Goodfellow \textit{et al.} 
in 2014 \cite{goodfellow2020generative}, have a variety of applications in media generation \cite{kwon2019predicting}, image restoration \cite{pan2021exploiting}, and data privacy \cite{jordon2018pate}. GANs aim to generate synthetic data that closely resembles the original real data with (unknown) underlying distribution $\Px$. The GAN is trained such that the distribution of the generated data, $\Pg$, approximates $\Px$ well. More specifically, low-dimensional random noise is fed to a generator neural network $G$ to produce synthetic data. Real data and the generated data are then given to a discriminator neural network $D$ scoring the data between 0 and 1, with a score close to 1 meaning that the discriminator thinks the data belongs to the real dataset. The discriminator and generator play a minimax game, where the aim is to minimize the generator's loss and maximize the discriminator's~loss.

Since their initial introduction, several variants of GAN have been proposed. Deep convolutional GAN (DCGAN) \cite{Radford2015} utilizes the same loss functions as VanillaGAN (the original GAN), combining GANs with convolutional neural networks, which are helpful when applying GANs to image data as they extract visual features from the data. DCGANs are more stable than the baseline model, but can suffer from mode collapse, which occurs when the generator learns that a select number of images can easily fool the discriminator, resulting in the generator only generating those images. Another notable issue with VanillaGAN is the tendency for the generator network's gradients to vanish. In the early stages of training, the discriminator lacks confidence, assigning generated data values close to zero. Therefore, the objective function tends to zero, resulting in small gradients and a lack of learning. To mitigate this issue, a non-saturating generator loss function was proposed in \cite{goodfellow2020generative} so that gradients do not vanish early on in training.

In the original (VanillaGAN) problem setup, the objective function, expressed as a negative sum of two Shannon cross-entropies, is to be minimized by the generator and maximized by the discriminator. It is demonstrated that if the discriminator is fixed to be optimal (i.e., as a maximizer of the objective function), the GAN's minimax game can be reduced to minimizing the Jensen-Shannon divergence (JSD) between the real and generated data's probability distributions \cite{goodfellow2020generative}. An analogous result was proven in \cite{Bhatia_2021} for R\'enyiGANs, a dual-objective GAN using distinct discriminator and generator loss functions. More specifically, under a canonical discriminator loss function (as in \cite{goodfellow2020generative}), and a generator loss function expressed in terms of two R\'enyi cross-entropies, it is shown that the R\'enyiGAN optimization problem reduces to minimizing the Jensen-R\'enyi divergence, hence extending VanillaGAN's result. 

Nowozin \textit{et al}.\ generalized VanillaGAN by formulating a class of loss functions in~\cite{nowozin2016fgan} parametrized by a lower semicontinuous convex function $f$, devising $f$-GAN. More specifically, the $f$-GAN problem consists of minimizing an $f$-divergence between the true data distribution and the generator distribution via a minimax optimization of a Fenchel conjugate representation of the $f$-divergence, where the VanillaGAN discriminator's role (as a binary classifier) is replaced by a variational function estimating the ratio of the true data and generator distributions.
The $f$-GAN loss function may be tedious to derive, as it requires the computation of the Fenchel conjugate of $f$. 
It can be shown that $f$-GAN can interpolate between VanillaGAN and HellingerGAN, among others \cite{nowozin2016fgan}. 

More recently, $\alpha$-GAN was presented in~\cite{sankar2021}, where the aim is to derive a class of loss functions parameterized by $\alpha > 0$, expressed in terms of a class probability estimation (CPE) loss between a real label $y \in \{0, 1\}$ and predicted label $\hat{y} \in [0, 1]$ \cite{sankar2021}. The ability to control $\alpha$ as a hyperparameter is beneficial to be able to apply one system to multiple datasets, as two datasets may be optimal under different $\alpha$ values. This work was further analyzed in~\cite{sankar2022} and expanded in~\cite{sankar2023} by introducing the dual-objective $(\alpha_D, \alpha_G)$-GAN, which allowed for the generator and discriminator loss functions to have a distinct $\alpha$~parameter with the aim of improving training stability. When $\alpha_D = \alpha_G$, the $\alpha$-GAN optimization reduces to minimizing an Arimoto divergence, as originally derived in~\cite{sankar2021}. Note that $\alpha$-GAN can recover several $f$-GANs, such as HellingerGAN, VanillaGAN, WassersteinGAN and Total Variation GAN \cite{sankar2021}.
Furthermore, in their more recent work which unifies \cite{sankar2021,sankar2022,sankar2023}, the authors establish, under some conditions, a one-to-one correspondence between CPE loss based GANs (such as $\alpha$-GANs) and $f$-GANs that use a symmetric $f$-divergence; see~\cite[Theorems~4-5 and~Corollary~1]{welfert2023addressing}. They also prove various generalization and estimation error bounds for $(\alpha_D, \alpha_G)$-GANs and illustrate their ability in mitigating training instability for synthetic Gaussian data as well as the Celeb-A and LSUN Classroom image datasets.
The various $(\alpha_D, \alpha_G)$-GAN equilibrium results do not provide an analogous result to the JSD and Jensen-R\'enyi divergence minimization for the VanillaGAN~\cite{goodfellow2020generative} and R\'enyiGAN~\cite{Bhatia_2021} problems, respectively, as it does not involve a Jensen-type divergence.\footnote{Given a divergence measure $\mathcal{D}(p\| q)$ between distributions $p$ and $q$ (i.,e., a positive-definite bivariate function: $\mathcal{D}(p\|q)\ge 0$ with equality if and only if (iff) $p=q$ almost everywhere (a.e.)), a {\em Jensen-type divergence of $\mathcal{D}$} is given by $\frac{1}{2}\mathcal{D}\big(p\|\frac{p+q}{2}\big)+\frac{1}{2}\mathcal{D}\big(q\|\frac{p+q}{2}\big)$; i.e., it is the arithmetic average of two $\mathcal{D}$-divergences, one between $p$ and the mixture $(p+q)/2$, and the other between $q$ and $(p+q)/2$.}

The main objective of our work is to present a unifying approach that provides an axiomatic framework to encompass several existing GAN generator loss functions so that the GAN optimization can be simplified in terms of a Jensen-type divergence. In particular, our framework classifies the set of $\alpha$-parameterized CPE-based loss functions $\La$, generalizing the $\alpha$-loss function in~\cite{sankar2021,sankar2022,sankar2023,welfert2023addressing}. We then propose $\La$-GAN, a dual objective GAN that uses a function from this class for the generator, and uses any canonical discriminator loss function that admits the same optimizer as VanillaGAN~\cite{goodfellow2020generative}. We show that under some regularity (convexity/concavity) conditions on $\La$, the minimax game played with these two loss functions is equivalent to the minimization of a Jensen-$f_\alpha$-divergence, a Jensen-type divergence and another natural extension of the Jensen-Shannon divergence (in addition to the Jensen-R\'enyi divergence~\cite{Bhatia_2021}), where the generating function $f_\alpha$ of the divergence is directly computed from the CPE loss function $\La$. This result recovers various prior dual-objective GAN equilibrium results, thus unifying them under one parameterized generator loss function. 
The newly obtained Jensen-$f_\alpha$-divergence, which is noted to belong to the class of symmetric $f$-divergences with different generating functions (see Remark~\ref{jensen-f-div-is-an-f-div}), 
is a useful measure of dissimilarity between distributions as it requires a convex function $f$ with a restricted domain given by the interval $[0,2]$ (see Remark~\ref{restricted-domain}) in addition to its symmetry and finiteness properties.

The rest of the paper is organized as follows. In Section~\ref{sec:preliminaries}, we review $f$-divergence measures and introduce the Jensen-$f$-divergence as an extension of the Jensen-Shannon divergence. In Section~\ref{sec:main_result}, we establish our main result regarding the optimization of our unifying generator loss function (Theorem~\ref{theorem:jensen-f}), and show that it can be applied to a large class of known GANs (Lemmas~\ref{lemma:vgan_example}-\ref{lemma:lkgan_jensen_f}). We conduct experiments in Section~\ref{sec:experiments} by implementing different manifestations of $\La$-GAN on three datasets, MNIST, CIFAR-10 and Stacked MNIST. Finally, we conclude the paper in Section~\ref{sec:conclusion}.

\section{Preliminaries}\label{sec:preliminaries}
We begin by presenting key information measures used throughout the paper. 
Let $f: [0, \infty) \to (-\infty,\infty]$ be a convex continuous function\footnote{The convexity of $f$ already implies its continuity on $(0,\infty)$. Here we extend the continuity of $f$ at~0, setting $f(0)= \lim_{u \downarrow 0} f(u)$, which may be infinite. Otherwise, $f(u)$ is assumed to be finite for $u>0$.} that is strictly convex at~1 (i.e., $f(\lambda u_1 + (1-\lambda)u_2) < \lambda f(u_1)+(1-\lambda) f(u_2)$ for all $u_1,u_2\ge 0$, $u_1\ne u_2$, and $\lambda \in (0,1)$ such that $\lambda u_1 + (1-\lambda)u_2=1$) and satisfying $f(1) = 0$.

\begin{definition}\textsc{\cite{csi,csiszar67,silvey}}\label{def:f_div} 
    The \textbf{$\boldsymbol{f}$-divergence} between two probability densities $p$ and $q$ with common support $\mathcal{R}\subseteq \mathbb{R}^d$ on the Lebesgue measurable space ($\mathcal{R}$, $\mathcal{B}(\mathcal{R})$, $\mu$) is denoted by $D_f(p\|q)$ and given~by\footnote{For simplicity, we consider throughout densities with common supports. A comprehensive definition of $f$-divergence for arbitrary distributions can be found in \cite[Section~III]{liese2006divergences}.}
    \begin{align}\label{eq:f_div}
        D_f(p\|q) &=  \intr{q\,f\left(\frac{p}{q}\right)},
    \end{align}
where we have used the shorthand $\intr{g} := \int_{\mathcal{R}}g(x)\,d\mu(x)$,  where $g$ is a measurable function; we follow this convention from now on. Here, $f$ is referred to as the generating function of $D_f(p\|q)$.
\end{definition}
We require that $f$ is strictly convex around~$1$ and that it satisfies the normalization condition $f(1) = 0$ to ensure positive-definiteness of the $f$-divergence, i.e., $D_f(p\|q) \ge 0$ with equality holding iff $p = q$ (a.e.). We present examples of $f$-divergences under various choices of their generating function~$f$ in Table~\ref{table:f_div_table}. We will be invoking these divergence measures in different parts of the paper. 
\begin{table}[H]
\caption{Examples of $f$-divergences.}
    \label{table:f_div_table}
    \resizebox{\textwidth}{!}{
    \centering

    \begin{tabular}{c c c c}
        \toprule
        $f$-Divergence & Symbol & Formula & $f(u)$ \\ \midrule
        Kullback-Leiber \cite{kl_div} & KL & $\intr{p\log\left(\frac{p}{q}\right)}$ & $u \log u$\\
        Jensen-Shannon \cite{nielsen2020generalization} & JSD & $\frac{1}{2}\kl\left(p\big{|}\big{|}\frac{p+q}{2}\right) + \frac{1}{2}\kl\left(q \big{|}\big{|} \frac{p+q}{2}\right)$ & $\frac{1}{2}\left( u \log u -(u+1)\log\frac{u+1}{2}\right)$\\
        Pearson $\chi^2$ \cite{nielsen2013chi} & $\chi^2$ & $\intr{\frac{(q-p)^2}{p}}$  & $\left(\sqrt{x}-\frac{1}{\sqrt{x}}\right)^2$\\
        Pearson-Vajda ($k > 1$) \cite{nielsen2013chi} & $|\chi|^k$ & $\intr{\frac{|q-p|^k}{p^{k-1}}}$ & $u^{1-k}|1-u|^k$\\
        Arimoto ($\alpha > 0$, $\alpha \neq 1$) \cite{ARIMOTO1971181,osterreicher1996class,liese2006divergences} & $\mathcal{A}_\alpha$ & $\frac{\alpha}{\alpha - 1}\left(\intr{(p^\alpha + q^\alpha)^{\frac{1}{\alpha}}} - 2^{\frac{1}{\alpha}}\right)$ & $\frac{\alpha}{\alpha - 1}\left((1+u)^{\frac{1}{\alpha}} - (1+u) - 2^{\frac{1}{\alpha}} + 2\right)$\\
        Hellinger ($\alpha > 0$, $\alpha \neq 1$) \cite{Hellinger+1909+210+271,liese2006divergences,Sason_2018} & $\mathscr{H}_\alpha$ & $\frac{1}{\alpha - 1}\left(\intr{p^{\alpha}q^{1-\alpha}} - 1\right)$ & $\frac{u^\alpha - 1}{\alpha - 1}$\\
        \bottomrule
    \end{tabular}
    }
\end{table}
The R\'enyi divergence of order $\alpha$ ($\alpha > 0$, $\alpha \neq 1$) between densities $p$ and $q$ with common support $\mathcal{R}$ is used in \cite{Bhatia_2021} in the R\'enyiGAN problem; it is given by~\cite{renyi1961measures,van2014renyi}
\begin{align}\label{eq:renyi_div}
    D_\alpha(p\|q) &= \frac{1}{\alpha - 1}\log\left(\intr{p^\alpha q^{1-\alpha}}\right).
\end{align}
Note that the R\'enyi divergence is not an $f$-divergence; however, it can be expressed as a transformation of the Hellinger divergence (which is itself an $f$-divergence):
\begin{align}
    D_\alpha(p \| q) &= \frac{1}{\alpha - 1}\log(1 + (\alpha - 1)\mathscr{H}_\alpha(p \| q)).
\end{align}
We now introduce a new measure, the Jensen-$f$-divergence, which is analogous to the Jensen-Shannon and Jensen-R\'enyi divergences.
\begin{definition}\label{def:jensen_f_div}
     The \textbf{Jensen-$\boldsymbol{f}$-divergence} between two probability distributions $p$ and $q$ with common support $\mathcal{R} \subseteq \mathbb{R}^d$ on the Lebesgue measurable space ($\mathcal{R}$, $\mathcal{B}(\mathcal{R})$, $\mu$) is denoted by $\jfd{f}(p\|q)$ and given by
\begin{align}
    \jfd{f}(p\|q) &= \frac{1}{2}D_f\left(p \bigg{|}\bigg{|} \frac{p+q}{2}\right) + \frac{1}{2}D_f\left(q \bigg{|}\bigg{|} \frac{p+q}{2}\right),\label{eq:jf_div}
\end{align}
where $D_f(\cdot \| \cdot)$ is the $f$-divergence.
\end{definition}

We next verify that the Jensen-Shannon divergence is a Jensen-$f$-divergence.
\begin{lemma}\label{lemma:jsd_as_jfd}
    Let $p$ and $q$ be two densities with common support $\mathcal{R} \subseteq \mathbb{R}^d$, and consider the function $f: [0, \infty) \to (-\infty,\infty]$ given by $f(u) = u\log u$. Then we have that
    \begin{align}
        \jfd{f}(p\|q) &= \textsc{JSD}(p\|q).
    \end{align}
\end{lemma}
\smallskip\noindent
{\bf Proof.}
    As $f$ is convex (and continuous) on its domain with $f(1)=0$, we have that
    \begin{align*}
        \textsc{JSD}(p\|q) &= \frac{1}{2}\kl\left(p\bigg{|}\bigg{|}\frac{p+q}{2}\right) + \frac{1}{2}\kl\left(q \bigg{|}\bigg{|} \frac{p+q}{2}\right)\\
    &= \frac{1}{2}\intr{p \log\left(\frac{2p}{p+q}\right)} + \frac{1}{2}\intr{q \log\left(\frac{2q}{p+q}\right)}\\
    &=\frac{1}{2}\intr{\frac{p+q}{2}\left(\frac{2p}{p+q}\log\left(\frac{2p}{p+q}\right)\right)} \\
    & \qquad + \frac{1}{2}\intr{\frac{p+q}{2}\left(\frac{2q}{p+q}\log\left(\frac{2q}{p+q}\right)\right)}\\
    &= \jfd{f}(p\|q).
    \end{align*}
    \qed

\begin{remark}
[Jensen-\boldmath{$f$}-divergence is a symmetric \boldmath{$f$}-divergence]
\label{jensen-f-div-is-an-f-div}
Note that $\jfd{f}(p\|q)$  is itself a symmetric $f$-divergence (with a modified generating function). Indeed, given the continuous convex function $f$ that is strictly convex around $1$ with $f(1)=0$, consider the functions
$$f_1(u) \coloneqq \frac{u+1}{2} \, f\Big(\frac{2u}{u+1}\Big), \qquad u\ge 0,$$
and
$$f_2(u) \coloneqq \frac{u+1}{2} \, f\Big(\frac{2}{u+1}\Big), \qquad u\ge 0,$$
which are both continuous convex, strictly convex around $1$, and satisfy $f_1(1)=f_2(1)=0$.
Now direct calculations yield that
$$D_f\left(p \bigg{|}\bigg{|} \frac{p+q}{2}\right)=D_{f_1}(p\|q)$$
and
$$D_f\left(q \bigg{|}\bigg{|} \frac{p+q}{2}\right)=D_{f_2}(p\|q).
$$
Thus
\begin{align*}
    \jfd{f}(p\|q) &= \frac{1}{2} D_{f_1}(p\|q) + \frac{1}{2} D_{f_2}(p\|q) 
   = D_{\bar{f}}(p\|q),
\end{align*}
where $\bar{f} \coloneqq \frac{1}{2}(f_1+f_2)$, i.e., 
\begin{equation}
\bar{f}(u)=\frac{u+1}{4} \Bigg( f\Big(\frac{2u}{u+1}\Big)+ f\Big(\frac{2}{u+1}\Big)\Bigg), \qquad u\ge 0,
\end{equation}
is also continuous convex, strictly convex around $1$ and satisfies $\bar{f}(1)=0$. Since by~\eqref{eq:jf_div},
$$\jfd{f}(p\|q)=\jfd{f}(q\|p),$$
we conclude that the Jensen-$f$-divergence is a symmetric $\bar{f}$-divergence.\footnote{Equivalently, we have that $\bar{f}=\bar{f}^\star$, where $\bar{f}^\star(u)\coloneqq u\bar{f}(\frac{1}{u})$, $u\ge 0$ (with $\bar{f}^\star(0)=\lim_{t \to \infty} \bar{f}(t)/t$), which is a necessary and sufficient condition for the $\bar{f}$-divergence to be symmetric~\cite[p.~4399]{liese2006divergences}.}
\end{remark}

\begin{remark}[\em Domain of \boldmath{$f$}]\label{restricted-domain}
Examining~\eqref{eq:jf_div}, we note that the Jensen-$f$-divergence between $p$ and $q$ involves the $f$-divergences between either $p$ or $q$ and their mixture $(p+q)/2$. In other words to determine $\jfd{f}(p\|q)$, we only need $f\big(\frac{2p}{p+q}\big)$ and $f\big(\frac{2q}{p+q}\big)$ when taking the expectations in~\eqref{eq:f_div}. Thus, it is sufficient to restrict the domain of the convex function $f$ to the interval $[0,2]$.
\end{remark}

\section{Main Results}\label{sec:main_result}
We now present our main theorem which unifies various generator loss functions under a CPE-based loss function $\La$ for a dual-objective GAN, $\La$-GAN, with a canonical discriminator loss function loss function that is optimized as in~\cite{goodfellow2020generative}. Under some regularity conditions on the loss-function $\La$, we show that under the optimal discriminator, our generator loss becomes a Jensen-$f$-divergence.

Let $(\mathcal{X}, \mathcal{B}(\mathcal{X}), \mu)$ be the measure space of $n \times n \times m$ images (where $m = 1$ for black and white images and $m = 3$ for RGB images), and let $(\mathcal{Z}, \mathcal{B}(\mathcal{Z}), \mu)$ be a measure space such that $\mathcal{Z} \subseteq \mathbb{R}^{d}$. The discriminator neural network is given by $D: \mathcal{X} \to [0, 1]$, and the generator neural network is given by $G: \mathcal{Z} \to \mathcal{X}$. The generator's noise input is sampled from a multivariate Gaussian distribution $\Pz: \mathcal{Z} \to [0, 1]$. We denote the probability distribution of the real data by $\Px: \mathcal{X} \to [0,1]$ and the probability distribution of the generated data by $\Pg: \mathcal{X} \to [0, 1]$. We also set $\Px$ and $\Pg$ as the densities corresponding to $\Px$ and $\Pg$, respectively. We begin by introducing the $\lagan$ system.
\begin{definition}\label{def:la_gan}
    Fix $\alpha \in \mathcal{A} \subseteq \mathbb{R}$ and let $\La: \{0, 1\} \times [0, 1] \to [0, \infty)$ be a loss function such that $\hat{y}\La\big(1,\frac{\hat{y}}{2}\big)$ is a continuous function that is either convex or concave in $\hat{y}\in [0,2]$, with strict convexity (resp., strict concavity) around $\hat{y}=1$,
    and such that $\La$ is symmetric in the sense that    \begin{align}\label{eq:symmetric_l}
        \La(1, \hat{y}) &= \La(0, 1-\hat{y}), \qquad \hat{y} \in [0, 1].
    \end{align}
    Then the $\boldlagan$ system is defined by $(V_D, V_{\La, G})$, where $V_D: \mathcal{X} \times \mathcal{Z} \to \mathbb{R}$ is the discriminator loss function, and $V_{\La, G}: \mathcal{X} \times \mathcal{Z} \to \mathbb{R}$ is the generator loss function, given by
    \begin{align}\label{eq:lagan_gen_loss}
        V_{\La, G}(D, G) &= \E_{\A \sim \Px}[-\La(1,D(\A))] + \E_{\B \sim \Pg}[-\La(0, D(\B))].
    \end{align}
    Moreover, the $\boldlagan$ \textbf{problem} is defined by
    \begin{align}
        &\sup_D\,V_D(D, G) \label{eq:lagan_dis_opt}\\
        &\inf_G\, V_{\La, G}(D, G).\label{eq:lagan_gen_opt}
    \end{align}
\end{definition}
We now present our main result about the $\lagan$ optimization problem.
\begin{theorem}\label{theorem:jensen-f}
For a fixed $\alpha \in \mathcal{A} \subseteq \mathbb{R}$ and $\La: \{0,1\} \times [0,1] \to [0, \infty)$, let $(V_D, V_{\La, G})$ be the loss functions of $\lagan$, and consider the joint optimization in \eqref{eq:lagan_dis_opt}-\eqref{eq:lagan_gen_opt}. If $V_D$ is a canonical loss function in the sense that it is maximized at $D = D^*$, where
\begin{align}\label{eq:optimal_discriminator}
    D^* = \frac{\Px}{\Px + \Pg},
\end{align}
then \eqref{eq:lagan_gen_opt} reduces to
\begin{align}\label{eq:main_result}
    \inf_G V_{\La, G}(D^*, G) &= \inf_G 2a \jfd{f_\alpha}(\Px\|\Pg) - 2ab,
\end{align}
where $\jfd{f_\alpha}(\cdot \| \cdot)$ is the Jensen-$f_\alpha$-divergence, and 
$f_\alpha: [0,2] \to \mathbb{R}$
is a continuous convex function, that is strictly convex around $1$, given by    \begin{align}\label{eq:f_to_l}
        f_\alpha(u) &=  -u\Big(\frac{1}{a}\La\Big(1, \frac{u}{2}\Big) - b\Big),
    \end{align} 
    where $a$ and $b$ are real constants chosen so that $f_\alpha(1) = 0$
    with $a<0$ (resp., $a>0$) if $u\La\big(1, \frac{u}{2}\big)$ is convex (resp., concave).
    Finally, \eqref{eq:main_result} is minimized when $\Px = \Pg$ (a.e.).
\end{theorem}
\smallskip\noindent
{\bf Proof.}
Under the assumption that $V_D$ is maximized at $D^* = \frac{\Px}{\Px + \Pg}$, we have that
\begin{align*}
    V_{\La, G}(D^*, G) &= \E_{\A \sim \Px}[-\La(1, D^*(\A))] + \E_{\B \sim \Pg}[-\La(0, D^*(\B))]\\
    &= -\int_\mathcal{X}\Px \La(1, D^*)\,d\mu - \int_\mathcal{X}\Pg \La(0, D^*)\,d\mu\\
    &= -\int_\mathcal{X}\Px \La\left(1, \frac{\Px}{\Px + \Pg}\right)\,d\mu -\int_\mathcal{X}\Pg\La\left(0, \frac{\Px}{\Px + \Pg}\right)\,d\mu\\
    &= -2\int_\mathcal{X}\left(\frac{\Px + \Pg}{2}\right)\frac{\Px}{\Px + \Pg}\La\left(1, \frac{\Px}{\Px + \Pg}\right)\,d\mu\\&\qquad -2\int_\mathcal{X}\left(\frac{\Px + \Pg}{2}\right)\frac{\Pg}{\Px + \Pg}\La\left(0, \frac{\Px}{\Px + \Pg}\right)\,d\mu\\
    &\overset{(a)}{=} -2\int_\mathcal{X}\left(\frac{\Px + \Pg}{2}\right)\frac{\Px}{\Px + \Pg}\La\left(1, \frac{\Px}{\Px + \Pg}\right)\,d\mu\\&\quad -2\int_\mathcal{X}\left(\frac{\Px + \Pg}{2}\right)\frac{\Pg}{\Px + \Pg}\La\left(1, \frac{\Pg}{\Px + \Pg}\right)\,d\mu\\
    &\overset{(b)}{=} -2\intx{\left(\frac{\Px+ \Pg}{2}\right)\frac{\Px}{\Px + \Pg}\left(\frac{-af_\alpha\left(\frac{2\Px}{\Px + \Pg}\right)}{\frac{2\Px}{\Px + \Pg}} + ab\right)}\\
    &\quad -2\intx{\left(\frac{\Px+ \Pg}{2}\right)\frac{\Pg}{\Px + \Pg}\left(\frac{-af_\alpha\left(\frac{2\Pg}{\Px + \Pg}\right)}{\frac{2\Pg}{\Px + \Pg}} + ab\right)}\\
    &= 2a\left(\frac{1}{2}\int_\mathcal{X}\frac{\Px + \Pg}{2}f_\alpha\left(\frac{2\Px}{\Px + \Pg}\right)\,d\mu\right.\\  
    &\quad +\left. \frac{1}{2}\int_\mathcal{X}\frac{\Px + \Pg}{2}f_\alpha\left(\frac{2\Pg}{\Px + \Pg}\right)\,d\mu \right) - 2ab\\
    &= 2a\, \jfd{f_\alpha}(\Px \| \Pg) - 2ab,
\end{align*}
where:
\begin{itemize}
    \item (a) holds since $\La(1, u) = \La(0, 1-u)$ by~\eqref{eq:symmetric_l}, where $u = \frac{\Px}{\Px + \Pg}$.
    \item (b) holds by solving for $\La(1, u)$ in terms of $f_\alpha(2u)$ in \eqref{eq:f_to_l}, where $u = \frac{\Px}{\Px + \Pg}$ in the first term and $u = \frac{\Pg}{\Px + \Pg}$ in the second term.
\end{itemize}
The constants $a$ and $b$ are chosen so that $f_\alpha(1)=0$. Finally, the continuity and convexity of $f_\alpha$ (as well as its strict convexity around $1$) directly follow from the corresponding assumptions imposed on the loss function~$\La$ in Definition~\ref{def:la_gan} and on the condition imposed on the sign of $a$ in the theorem's statement.
\qed
\begin{remark}
    Note that not only $D^*$ given in \eqref{eq:optimal_discriminator} is an optimal discriminator of the (original) VanillaGAN discriminator loss function, but it also optimizes the LSGAN/L$k$GAN discriminator loss function when their discriminator's labels for fake and real data, $\gamma$ and $\beta$, respectively satisfy $\gamma = 1$ and $\beta = 0$ (see Section \ref{subsec:lkgan}).
\end{remark}
We next show that the $\lagan$ of Theorem \ref{theorem:jensen-f} recovers as special cases a number of well-known GAN generator loss functions and their equilibrium points (under an optimal classical discriminator $D^*$).
\subsection{VanillaGAN}\label{subsec:vanillagan}
VanillaGAN \cite{goodfellow2020generative} uses the same loss function $\Vvg$ for both generator and discriminator, which is
\begin{align}
    \label{eq:vgan_loss}
    \Vvg(D, G) &= \E_{\A \sim \Px}[-\log D(\A)] + \E_{\B \sim \Pg}[-\log(1-D(\B))],
\end{align}
and can be cast as a saddle point optimization problem:
\begin{align}\label{eq:vgan_opt}
    \inf_G \sup_D \Vvg(D, G).
\end{align}
It is shown in \cite{goodfellow2020generative} that the optimal discriminator for \eqref{eq:vgan_opt} is given by $D^* = \frac{\Px}{\Px + \Pg}$, as in \eqref{eq:optimal_discriminator}.
When $D = D^*$, the optimization reduces to minimizing the Jensen-Shannon divergence:
\begin{align}\label{eq:vgan_sol}
    \inf_G\, \Vvg(D^*, G) &= \inf_G\, 2\textsc{JSD}(\Px \| \Pg) - 2\log{2}.
\end{align}
We next show that \eqref{eq:vgan_sol} can be obtained from Theorem \ref{theorem:jensen-f}.
\begin{lemma}\label{lemma:vgan_example}
Consider the optimization of the VanillaGAN given in \eqref{eq:vgan_opt}. Then we have that
\begin{align*}
   \Vvg(D^*, G) &= 2\textsc{JSD}(\Px \| \Pg) - 2\log 2 = V_{\La, G}(D^*, G),
\end{align*}
where $\La(y, \hat{y}) = -y\log(\hat{y}) - (1-y)\log(1-\hat{y})$ for all $\alpha \in \mathcal{A} = \mathbb{R}$.
\end{lemma}
\smallskip\noindent
{\bf Proof.}
For any fixed $\alpha \in \mathbb{R}$, let the function $\La$ in \eqref{eq:lagan_gen_loss} be as defined in the statement:
\begin{align*}
    \La(y, \hat{y}) &= -y\log(\hat{y}) - (1-y)\log(1-\hat{y}).
\end{align*}
Note that $\La$ is symmetric, since for $\hat{y} \in [0, 1]$, we have that
\begin{align*}
    \La(1, \hat{y}) = -\log(\hat{y}) = \La(0, 1-\hat{y}).
\end{align*}
Instead of showing the continuity and convexity/concavity conditions imposed on 
$\hat{y}\La\big(1,\frac{\hat{y}}{2}\big)$  in Definition~\ref{def:la_gan},
we implicitly verify them by directly deriving $f_\alpha$ from $\La$ using~\eqref{eq:f_to_l}
and showing that it is continuous convex and strictly convex around $1$.
Setting $a = 1$ and $b = \log 2$, we have that
\begin{align*}
    f_\alpha(u) &= -u\left(\frac{1}{a}\La\left(1, \frac{u}{2}\right) - b\right) \\
    & = -u\left(-\log\frac{u}{2} - \log 2\right) 
    = u \log u.
\end{align*}
Clearly, $f$ is convex (actually strictly convex on $(0,\infty)$ and hence strictly convex around $1$) and~continuous on its domain (where $f(0)=\lim_{u \to 0} u \log(u)=0$). It also satisfies $f(1) = 0$.
By Lemma~\ref{lemma:jsd_as_jfd}, we know that under the generating function $f(u) = u\log(u)$, the Jensen-$f$ divergence reduces to the Jensen-Shannon divergence.
Therefore, by Theorem \ref{theorem:jensen-f}, we have that
\begin{align*}
    V_{\La, G}(D^*, G) &= 2a \jfd{f_\alpha}(\Px \| \Pg) - 2ab \\
    &= 2\textsc{JSD}(\Px \| \Pg) - 2 \log 2 \\
    &= \Vvg(D^*, G),
\end{align*}
which finishes the proof. 
\qed

\subsection{\boldmath{$\alpha$}-GAN}\label{subsec:alphagan}
The notion of $\alpha$-GANs is introduced in \cite{sankar2021} as a way to unify several existing GANs using a parameterized loss function. We describe $\alpha$-GANs next.
\begin{definition}\label{def:alphaloss}\textsc{\cite{sankar2021}}
    Let $y \in \{0, 1\}$ be a binary label, $\hat{y} \in [0, 1]$, and fix $\alpha > 0$. The \textbf{$\boldsymbol{\alpha}$-loss} between $y$ and $\hat{y}$ is the map $\ell_\alpha: \{0, 1\} \times [0, 1] \to [0,\infty)$ given by
    \begin{align}
        \ell_\alpha(y, \hat{y}) &= \begin{dcases}
            \frac{\alpha}{\alpha - 1}\left(1-y\hat{y}^{\frac{\alpha-1}{\alpha}} + (1-y)(1-\hat{y})^{\frac{\alpha - 1}{\alpha}}\right), & \alpha \in (0, 1) \cup (1, \infty)\\
            -y\log\hat{y} - (1-y)\log(1-\hat{y}), & \alpha = 1.
        \end{dcases}
    \end{align}
\end{definition}
\begin{definition}\label{def:alphagan1}\textsc{\cite{sankar2021}}
    For $\alpha > 0$, the $\boldagan$\textbf{ loss function} is given by
    \begin{align}\label{eq:alphagan_loss1}
        V_\alpha(D, G) &= \E_{\A \sim \Px}[-\ell_\alpha(1, D(\A))] + \E_{\B \sim \Pg}[-\ell_\alpha(0, D(\B))].
    \end{align}
    The joint optimization of the $\agan$ problem is given by
    \begin{align}\label{eq:alphagan_opt1}
        \inf_G\,\sup_D\, V_\alpha(D, G).
    \end{align}
\end{definition}
It is known that $\alpha$-GAN recovers several well-known GANs by varying the $\alpha$ parameter, notably, the VanillaGAN ($\alpha = 1$) \cite{goodfellow2020generative} and the HellingerGAN ($\alpha = \frac{1}{2}$) \cite{nowozin2016fgan}. Furthermore, as $\alpha \to \infty$, $V_\alpha$ recovers a translated version of the WassersteinGAN loss function \cite{arjovsky2017wasserstein}. We now present the solution to the joint optimization problem presented in \eqref{eq:alphagan_opt1}.
\begin{proposition}\label{prop:alphagan1_sol}\textsc{\cite{sankar2021}}
    Let $\alpha > 0$, and consider the joint optimization of the $\alpha$-GAN presented in \eqref{eq:alphagan_opt1}. The discriminator $D^*$ that maximizes the loss function is given by
    \begin{align}\label{eq:alphagan1_opt_dis}
        D^* &= \frac{\Px^\alpha}{\Px^\alpha + \Pg^\alpha}.
    \end{align}
    Furthermore, when $D = D^*$ is fixed, the problem in \eqref{eq:alphagan_opt1} reduces to minimizing an Arimoto divergence (as defined in Table \ref{table:f_div_table}) when $\alpha \neq 1$:
    \begin{align}\label{eq:alphagan1_sol1}
        \inf_G V_\alpha(D^*, G) &= \inf_G \mathcal{A}_\alpha(\Px \| \Pg) + \frac{\alpha}{\alpha - 1}\left(2^{\frac{1}{\alpha}} -2 \right),
    \end{align}
    and a Jensen-Shannon divergence when $\alpha = 1$:
        \begin{align}\label{eq:alphagan1_sol2}
            \inf_G V_1(D^*, G) &= \inf_G \textsc{JSD}(\Px \| \Pg) - 2\log 2,
        \end{align}
    where \eqref{eq:alphagan1_sol1} and \eqref{eq:alphagan1_sol2} achieve their minima iff $\Px = \Pg$ (a.e.).
\end{proposition}

Recently, $\alpha$-GAN was generalized in~\cite{sankar2023} to implement a dual objective GAN, which we describe~next.
\begin{definition}\label{def:alphagan2}\textsc{\cite{sankar2023}}
    For $\alpha_D>0$ and $\alpha_G > 0$, the $\boldadggan$\textbf{'s optimization} is given by
    \begin{align}
        &\sup_D\, V_{\alpha_D}(D, G)\label{eq:alphagan2_sup_dis}  \\
        &\inf_G\,V_{\alpha_G}(D, G)\label{eq:alphagan2_inf_gen}
    \end{align}
    where $V_{\alpha_D}$ and $V_{\alpha_G}$ are defined in \eqref{eq:alphagan_loss1}, with $\alpha$ replaced by $\alpha_D$ and $\alpha_G$ respectively.
\end{definition}
\begin{proposition}\label{prop:alphagan2_sol}\textsc{\cite{sankar2023}}
Consider the joint optimization in \eqref{eq:alphagan2_sup_dis}-\eqref{eq:alphagan2_inf_gen}. Let parameters $\alpha_D$, $\alpha_G > 0$ satisfy
\begin{align}
    \left(\alpha_D \leq 1, \alpha_G > \frac{\alpha_D}{\alpha_D + 1}\right) \text{ or } \left(\alpha_D > 1, \frac{\alpha_D}{2} < \alpha_G \leq \alpha_D\right).
\end{align}
The discriminator $D^*$ that maximizes $V_{\alpha_D}$ is given by
\begin{align}\label{eq:alphagan2_opt_dis}
    D^* &= \frac{\Px^{\alpha_D}}{\Px^{\alpha_D} + \Pg^{\alpha_D}}.
\end{align}
Furthermore, when $D = D^*$ is fixed, the minimization of $V_{\alpha_G}$ in \eqref{eq:alphagan2_inf_gen} is equivalent to the following $f$-divergence minimization:
\begin{align}\label{eq:alphagan2_sol}
    \inf_G\,V_{\alpha_G}(D^*, G) &= \inf_G\,D_{f_{\alpha_D, \alpha_G}}(\Px \| \Pg) + \frac{\alpha}{\alpha - 1}\left(2^{\frac{1}{\alpha}} - 2\right),
\end{align}
where $f_{\alpha_D, \alpha_G}: [0, \infty) \to \mathbb{R}$ is given by
\begin{align}\label{eq:alphagan2_f_div_function}
    f_{\alpha_D, \alpha_G}(u) &= \frac{\alpha_G}{\alpha_G - 1}\left(\frac{u^{\alpha_D\left(1-\frac{1}{\alpha_G}\right) + 1} + 1}{(u^{\alpha_D} + 1)^{1-\frac{1}{\alpha_G}}}\right).
\end{align}
\end{proposition}
We now apply the $(\alpha_D, \alpha_G)$-GAN to our main result in Theorem \ref{theorem:jensen-f} by showing that \eqref{eq:main_result} can recover \eqref{eq:alphagan2_sol} when $\alpha_D = 1$ (which corresponds to a VanillaGAN discriminator loss function).
\begin{lemma}\label{lemma:alphagan_example}
Consider the $\adggan$ given in Definition \ref{def:alphagan2}. Let $\alpha_D = 1$ and $\alpha_G = \alpha > \frac{1}{2}$. Then, the solution to \eqref{eq:alphagan2_inf_gen} presented in Proposition \ref{prop:alphagan2_sol} is equivalent to minimizing a Jensen-$f_\alpha$-divergence: specifically, if $D^*$ is the optimal discriminator given by \eqref{eq:alphagan2_opt_dis}, which is equivalent to \eqref{eq:optimal_discriminator} when $\alpha_D = 1$, then $V_{\alpha, G}(D^*, G)$ in \eqref{eq:alphagan2_sol} satisfies
\begin{align}\label{eq:alphagan_jensen_f_opt}
    V_{\alpha,G}(D^*, G) &= 2^{\frac{1}{\alpha}}\jfd{f_\alpha}(\Px \| \Pg) + \frac{\alpha}{\alpha - 1}(2^\frac{1}{\alpha} - 2) = V_{\La, G}(D^*, G),
\end{align}
where $\La(y, \hat{y}) = \ell_\alpha(y, \hat{y})$ and
\begin{align}
    f_\alpha(u) = \frac{\alpha}{\alpha - 1}\left(u^{2-\frac{1}{\alpha}} - u\right), \quad u \geq 0.
\end{align}
\end{lemma}
\smallskip\noindent
{\bf Proof.}
We show that Theorem \ref{theorem:jensen-f} recovers Proposition~\ref{prop:alphagan2_sol} by setting $\La(y,\hat{y}) = \ell_\alpha(y, \hat{y})$. Note that $\ell_\alpha$ is symmetric, since
\begin{align*}
    \ell_\alpha(1, \hat{y}) = \frac{\alpha}{\alpha - 1}(1- \hat{y}^{1-\frac{1}{\alpha}}) = \ell_\alpha(0, 1-\hat{y}).
\end{align*}
As in the proof of Lemma~\ref{lemma:vgan_example}, instead of proving the conditions imposed on 
$\hat{y}\La\big(1,\frac{\hat{y}}{2}\big)$  in Definition~\ref{def:la_gan},
we derive $f_\alpha$ directly from $\La$ using~\eqref{eq:f_to_l} and show that it is continuous convex and strictly convex around $1$.
From Lemma~\ref{lemma:vgan_example}, we know that when $\alpha = 1$, $f_\alpha(u) = u\log u$ (which is strictly convex and continuous). For $\alpha \in (0, 1) \cup (1, \infty)$, setting 
$a = 2^{\frac{1}{\alpha} - 1}$
and~$b = \frac{\alpha}{\alpha - 1}\left(2^{1-\frac{1}{\alpha}} - 1\right)$
in~\eqref{eq:f_to_l}, we have that
\begin{align*}
    f_\alpha(u) &= -u\left(\frac{1}{a}\La\left(1, \frac{u}{2}\right) - b\right)\\
    &= -u\left(2^{1-\frac{1}{\alpha}}\frac{\alpha}{\alpha - 1}\left(1-\left(\frac{u}{2}\right)^{1-\frac{1}{\alpha}}\right) - \frac{\alpha}{\alpha - 1}(2^{1-\frac{1}{\alpha}} - 1)\right)\\
    &= \frac{\alpha}{\alpha - 1}(-u)[2^{1-\frac{1}{\alpha}} - u^{1-\frac{1}{\alpha}} - (2^{1-\frac{1}{\alpha}} - 1)]\\
    &= \frac{\alpha}{\alpha - 1}(u^{2-\frac{1}{\alpha}} - u).
\end{align*}
Clearly $f_\alpha(1) = 0$. Furthermore for $\alpha \neq 1$, we have that
\begin{align*}
    f_\alpha^{\prime \prime}(u) &= \frac{(2\alpha - 1)u^{\frac{-1}{\alpha}}}{\alpha},\quad u \ge 0,
\end{align*}
which is positive for $\alpha > \frac{1}{2}$, and $f_\alpha$ is convex for $\alpha > \frac{1}{2}$ (as well as continuous on its domain and strictly convex around $1$). Thus by Theorem~\ref{theorem:jensen-f}, we have that
\begin{align*}
    V_{\La, G}(D^*, G) &= 2a \jfd{f_\alpha}(\Px \| \Pg) - 2ab\\
    &= 2\cdot2^{\frac{1}{\alpha} - 1}\jfd{f_\alpha}(\Px \| \Pg) - 2\frac{\alpha}{\alpha - 1}2^{\frac{1}{\alpha} - 1}(2^{1 - \frac{1}{\alpha}} - 1)\\
    &= 2^{\frac{1}{\alpha}}\jfd{f_\alpha}(\Px \| \Pg) + \frac{\alpha}{\alpha - 1}(2^\frac{1}{\alpha} - 2).
\end{align*}
We now show that the above Jensen-$f_\alpha$-divergence is equal to the $f_{1, \alpha}$-divergence originally derived for the $(1, \alpha)$-GAN problem of Proposition~\ref{prop:alphagan2_sol} (note from Proposition~\ref{prop:alphagan2_sol}, that if $\alpha_D = 1$, then $\alpha_G = \alpha > \frac{1}{2}$, so the range of $\alpha$ concurs with the range above required for the convexity of $f_\alpha$). For any two distributions $p$ and $q$ with common support $\mathcal{X}$, we have that
\begin{align*}
    D_{f_{1, \alpha}}(p\|q) &= \frac{\alpha}{\alpha - 1}\intx{q\frac{\left(\frac{p}{q}\right)^{2-\frac{1}{\alpha}}+1}{\left(\frac{p}{q} + 1\right)^{1- \frac{1}{\alpha}}}} - \frac{\alpha}{\alpha - 1}2^{\frac{1}{\alpha}}\\
    &= \frac{\alpha}{\alpha - 1}\intx{q\frac{\left(\frac{p}{q}\right)^{2- \frac{1}{\alpha}} + 1}{\left(\frac{p+q}{q}\right)^{1-\frac{1}{\alpha}}}} - \frac{\alpha}{\alpha-1}2^{\frac{1}{\alpha}}\\
    &= \frac{\alpha}{\alpha - 1}\intx{\left((p+q)\left(\frac{p}{p+q}\right)^{2-\frac{1}{\alpha}} + (p+q)\left(\frac{q}{p+q}\right)^{2-\frac{1}{\alpha}}\right)}\\ &\qquad- \frac{\alpha}{\alpha - 1}2^{\frac{1}{\alpha}}\\
   &= \frac{\alpha}{\alpha - 1}\frac{2}{2^{2-\frac{1}{\alpha}}}\intx{\left(\frac{p+q}{2}\left(\frac{2p}{p+q}\right)^{2-\frac{1}{\alpha}} + \frac{p+q}{2}\left(\frac{2q}{p+q}\right)^{2-\frac{1}{\alpha}}\right)}\\&\qquad - \frac{\alpha}{\alpha - 1}2^{\frac{1}{\alpha}}\\
   &= \frac{\alpha}{\alpha - 1}2^{\frac{1}{\alpha} - 1}\intx{\left(\frac{p+q}{2}\left(\left(\frac{2p}{p+q}\right)^{2-\frac{1}{\alpha}}-\frac{2p}{p+q}\right) + p\right)} \\
   &\qquad+\frac{\alpha}{\alpha - 1}2^{\frac{1}{\alpha} - 1}\intx{\left(\frac{p+q}{2}\left(\left(\frac{2q}{p+q}\right)^{2-\frac{1}{\alpha}}-\frac{2q}{p+q}\right) + q\right)}\\&\quad\qquad - \frac{\alpha}{\alpha-1}2^{\frac{1}{\alpha}}\\
   &= \frac{\alpha}{\alpha - 1}2^{\frac{1}{\alpha}}\frac{1}{2}\left(\intx{\frac{p+q}{2}\left(\left(\frac{2p}{p+q}\right)^{2-\frac{1}{\alpha}} - \frac{2p}{p+q}\right)} + 1 \right)\\
   &\qquad+\frac{\alpha}{\alpha - 1}2^{\frac{1}{\alpha}}\frac{1}{2}\left(\intx{\frac{p+q}{2}\left(\left(\frac{2q}{p+q}\right)^{2-\frac{1}{\alpha}} - \frac{2q}{p+q}\right)} + 1 \right)\\&\quad\qquad - \frac{\alpha}{\alpha - 1}2^{\frac{1}{\alpha}}\\
   &= 2^{\frac{1}{\alpha}}\jfd{f_\alpha}(p\|q) + \frac{\alpha}{\alpha - 1}2^{\frac{1}{\alpha} - 1}(2) - \frac{\alpha}{\alpha - 1}2^{\frac{1}{\alpha}}\\
   &= 2^{\frac{1}{\alpha}}\jfd{f_\alpha}(p\|q).
\end{align*}
Therefore, $V_{\La, G}(D^*, G) = V_\alpha(D^*, G)$.
\qed

Note that this lemma generalizes Lemma \ref{lemma:vgan_example}; the VanillaGAN is a special case of the $(1, \alpha)$-GAN for $\alpha = 1$.
\subsection{Shifted LkGANs and LSGANs}\label{subsec:lkgan}
Least Squares GAN (LSGAN) was proposed in \cite{mao2017squares} to mitigate the vanishing gradient problem with VanillaGAN and to stabilize training performance. The LSGAN's loss function is derived from the squared error distortion measure, where we aim to minimize the distortion between the data samples and a target value we want the discriminator to assign the samples to. The LSGAN was generalized with the L$k$GAN in \cite{Bhatia_2021} by replacing the squared error distortion measure with the absolute error distortion measure of order $k \geq 1$, therefore introducing an additional degree of freedom to the generator's loss function. We first state the general L$k$GAN problem. We then apply the result of Theorem \ref{theorem:jensen-f} to the loss functions of LSGAN and L$k$GAN.
\begin{definition}\label{def:lkgan}\textsc{\cite{Bhatia_2021}}
Let $\gamma$, $\beta$, $c \in [0,1]$ and let $k \geq 1$. The \textbf{L$\boldsymbol{k}$GAN's loss functions}, denoted by $V_{\text{\tiny{LSGAN}}, D}$ and $V_{k, G}$ are given by
\begin{align}\label{eq:lkgan_dis_loss}
    V_{\text{\tiny{LSGAN}}, D}(D, G) &= -\frac{1}{2}\E_{\A \sim \Px}[(D(\A) - \beta)^2] - \frac{1}{2}\E_{\B \sim \Pg}[(D(\B) - \gamma)^2]
\end{align}
\begin{align}\label{eq:lkgan_gen_loss}
    V_{k, G}(D, G) &= \E_{\A \sim \Px}[|D(\A) - c|^k] + \E_{\B \sim \Pg}[|D(\B) - c|^k].
\end{align}
The \textbf{L$\boldsymbol{k}$GAN problem} is the joint optimization
\begin{align}\label{eq:lkgan_opt}
    &\sup_{D}\, V_{\text{\tiny{LSGAN}}, D}(D, G)\\
    &\inf_G\, V_{k, G}(D, G).
\end{align}
\end{definition}
We next recall the solution to \eqref{eq:lkgan_opt}, which is a minimization of the Pearson-Vajda divergence $|\chi|^k(\cdot \| \cdot)$ of order $k$ (as defined in Table \ref{table:f_div_table}).
\begin{proposition}\label{prop:lkgan_sol}\textsc{\cite{Bhatia_2021}}
    Consider the joint optimization for the L$k$GAN presented in \eqref{eq:lkgan_opt}. Then, the optimal discriminator $D^*$ that maximizes $V_{\text{\tiny{LSGAN}}, D}$ in \eqref{eq:lkgan_dis_loss} is given by
    \begin{align}\label{eq:lkgan_opt_dis}
        D^* &= \frac{\gamma\Px + \beta \Pg}{\Px + \Pg}.
    \end{align}
    Furthermore, if $D = D^*$, and $\gamma - \beta = 2(c - \beta)$, the minimization of $V_{k,G}$ in \eqref{eq:lkgan_gen_loss} reduces to
    \begin{align}
        \inf_G\,V_{k, G}(D, G) &= \inf_G\, |c - \beta|^k |\chi|^k(\Px + \Pg \| 2\Pg).
    \end{align}
\end{proposition}
Note that the LSGAN \cite{mao2017squares} is a special case of L$k$GAN, as we recover LSGAN when $k = 2$ \cite{Bhatia_2021}.

By scrutinizing Proposition~\ref{prop:lkgan_sol} and Theorem \ref{theorem:jensen-f}, we observe that the former cannot be recovered from the latter. However we can use Theorem \ref{theorem:jensen-f} by slightly modifying the L$k$GAN generator's loss function. First, for the dual objective GAN proposed in Theorem \ref{theorem:jensen-f}, we need $D^* = \frac{\Px}{\Px + \Pg}$. By \eqref{eq:lkgan_opt_dis}, this is achieved for $\gamma = 1$ and $\beta = 0$. Then, we define the intermediate loss function
\begin{align}\label{eq:lkgan_gen_loss2}
    \tilde{V}_{k, G}(D, G) &= \E_{\A \sim \Px}[|D(\A) - c_1|^k] + \E_{\B \sim \Pg}[|D(\B) - c_2|^k].
\end{align} 
Comparing the above loss function with \eqref{eq:lagan_gen_loss}, we note that setting $c_1 = 0$ and $c_2 = 1$ in \eqref{eq:lkgan_gen_loss2} satisfies the symmetry property of $\La$.
Finally, to ensure the generating function $f_\alpha$ satisfies $f_\alpha(1) = 0$, we shift each term in \eqref{eq:lkgan_gen_loss2} by 1. Putting these changes together, we propose a revised generator loss function, denoted by $\hat{V}_{k, G}$, given by
\begin{align}\label{eq:lkgan_gen_loss3}
    \hat{V}_{k, G}(D, G) &= \E_{\A \sim \Px}[|D(\A)|^k - 1] + \E_{\B \sim \Pg}[|1 - D(\B) |^k - 1].
\end{align}
We call a system that uses \eqref{eq:lkgan_gen_loss3} as a generator loss function a \textbf{Shifted L$\bm{k}$GAN (SL$\boldsymbol{k}$GAN)}. If $k = 2$, we have a shifted version of the LSGAN generator loss function, which we call the \textbf{Shifted LSGAN (SLSGAN)}. Note that none of these modifications alter the gradients of $V_{k, G}$ in \eqref{eq:lkgan_gen_loss}, since the first term is independent of $G$, the choice of $c_1$ is irrelevant, and translating a function by a constant does not change its gradients. However, from Proposition \ref{prop:lkgan_sol}, for $\gamma = 0$, $\beta = 1$ and $c = 1$, we do not have that $\gamma - \beta = 2(c - \beta)$, and as a result, this modified problem does not reduce to minimizing a Pearson-Vajda divergence. Consequently, we can relax the condition on $k$ in Definition \ref{def:lkgan} to just $k > 0$. We now show how Theorem \ref{theorem:jensen-f} can be applied to $\La$-GAN using \eqref{eq:lkgan_gen_loss3}.

\begin{lemma}\label{lemma:lkgan_jensen_f}
    Let $k > 0$. Let $V_D$ be a discriminator loss function, and let $\hat{V}_{k,G}$ be the generator's loss function defined in (\ref{eq:lkgan_gen_loss3}). Consider the joint optimization
\begin{align}\label{eq:lkgan_opt_v2}
    &\sup_D V_D(D, G)\\
    &\inf_G \hat{V}_{k, G}(D,G)
\end{align}
If $V_D$ is optimized at $D^* = \frac{\Px}{\Px + \Pg}$ (i.e., $V_D$ is canonical), then we have that
\begin{align*}
\hat{V}_{k, G}(D^*, G) &= \frac{1}{2^{k-1}}\jfd{f_k}(\Px \| \Pg) + \frac{1}{2^{k-1}} - \frac{1}{2},
\end{align*}
where $f_k$ is given by
\begin{align*}
    f_k(u) &= u(u^k-1), \qquad u\ge 0.
\end{align*}
\end{lemma}
Examples of $V_D(D, G)$ that satisfy the requirements of Lemma \ref{lemma:lkgan_jensen_f} include the L$k$GAN discriminator loss function given by \eqref{eq:lkgan_dis_loss} with $\gamma = 1$ and $\beta = 0$, and the VanillaGAN discriminator loss function given by \eqref{eq:vgan_loss}.

\smallskip\noindent
{\bf Proof.}
Let $k > 0$. We can restate the SL$k$GAN's generator loss function in~\eqref{eq:lkgan_gen_loss3} in terms of $V_{\La, G}$ in \eqref{eq:lagan_gen_loss}: we have that $V_{\La, G}(D^*, G) = \hat{V}_{k, G}(D^*, G)$, where $\alpha = k$ and $\Lk: \{0, 1\} \times [0, 1] \to [0,\infty)$ is given by
\begin{align}
    \Lk(y, \hat{y}) &= -(y(\hat{y}^k - 1) + (1-y)((1-\hat{y})^k - 1)).
\end{align}
We have that $\Lk$ is symmetric, since 
\begin{align*}
    \Lk(1, \hat{y}) &= -(\hat{y}^k - 1) 
    = \Lk(0, 1-\hat{y}).
\end{align*}
We derive $f_\alpha$ from $\La$ via~\eqref{eq:f_to_l} and
directly check that it is continuous convex and strictly convex around $1$.
Setting $a = \frac{1}{2^k}$ and $b = 2^k - 1$ in \eqref{eq:f_to_l}, we have that
\begin{align*}
    f_k(u) &= -u\left(\frac{1}{a}\Lk\left(1, \frac{u}{2}\right) - b\right)\\
    &= -u\left(2^{k}\left(1-\left(\frac{u}{2}\right)^k\right) - (2^k - 1)\right)\\
    &= -u(2^k - u^k - 2^k + 1)\\
    &= u(u^k - 1).
\end{align*}
We clearly have that $f_k(1) = 0$ and that $f_k$ is continuous. Furthermore, we have that $f_k^{\prime \prime}(u) = k(k+1)u$, which is non-negative for $u \ge 0$. Therefore $f_k$ is convex (as well as strictly convex around $1$). As a result, by Theorem \ref{theorem:jensen-f}, we have that
\begin{align*}
    \hat{V}_{k, G}(D^*, G) &= \frac{1}{2^{k-1}}\jfd{f_k}(\Px \| \Pg) - \frac{1}{2^{k-1}}(2^{k} -1)\\
    &= \frac{1}{2^{k-1}}\jfd{f_k}(\Px \| \Pg) + \frac{1}{2^{k-1}} - \frac{1}{2}.
\end{align*}
\qed

We conclude this section by emphasizing that Theorem~\ref{theorem:jensen-f} serves as a unifying result recovering the existing loss functions in the literature and moreover, provides a way for generalizing new ones. Our aim in the next section is to demonstrate the versatility of this result in experimentation.

\section{Experiments}\label{sec:experiments}

We perform two experiments on three different image datasets which we describe below. 
\smallskip

\noindent{\bf Experiment~1.} In the first experiment, we compare the $(\alpha, \alpha)$-GAN with the $(1, \alpha)$-GAN, controlling the value of $\alpha$.\footnote{We herein confine the comparison of $(1, \alpha)$-GAN with $(\alpha, \alpha)$-GAN only so that both systems have the same tunable free parameter $\alpha$. Results obtained in~\cite{sankar2023} for the Stacked MNIST dataset show that $(\alpha_D, \alpha_G)$-GAN provides a consistently robust performance when $\alpha_D=\alpha_G$.
Other experiments illustrating the performance of $(\alpha_D, \alpha_G)$-GAN with $\alpha_D \ne 1$ are carried for the Celeb-A and LSUN Classroom image datasets in~\cite{welfert2023addressing}, showing improved training stability for $\alpha_D<1$ values.}
Recall that $\alpha_D = 1$ corresponds to the canonical VanillaGAN (or DCGAN) discriminator. We aim to verify whether or not replacing an $\alpha$-GAN discriminator with a VanillaGAN discriminator stabilizes or improves the system's performance depending on the value of $\alpha$. Note that the result of Theorem~\ref{theorem:jensen-f} only applies to the $(\alpha_D, \alpha_G)$-GAN for $\alpha_D = 1$. 

\medskip
\noindent{\bf Experiment~2.} We train two variants of SL$k$GAN, with the generator loss function as described in~\eqref{eq:lkgan_gen_loss3}, parameterized by $k > 0$. We then utilize two different canonical discriminator loss functions to align with Theorem~\ref{theorem:jensen-f}. The first is the VanillaGAN discriminator loss given by \eqref{eq:vgan_loss}; we call the resulting dual objective GAN by \textbf{Vanilla-SL$\boldsymbol{k}$GAN}. The second is the L$k$GAN discriminator loss, given by \eqref{eq:lkgan_dis_loss}, where we set $\gamma = 1$ and $\beta = 0$ such that the optimal discriminator is given by \eqref{eq:optimal_discriminator}. We call this system by \textbf{L$\boldsymbol{k}$-SL$\boldsymbol{k}$GAN}. We compare the two variants to analyze how the value of $k$ and choice of discriminator loss impacts the system's performance.

\subsection{Experimental Setup}
We run both experiments on three image datasets: MNIST \cite{mnist}, CIFAR-10 \cite{cifar10}, and Stacked MNIST \cite{NEURIPS2018_288cc0ff}. The MNIST dataset is a dataset of black and white handwritten digits between 0 and 9 of size $28 \times 28 \times 1$. The CIFAR-10 dataset is an RGB dataset of small images of common animals and modes of transportation of size $32 \times 32 \times 3$. The Stacked MNIST dataset is an RGB dataset derived from the MNIST dataset, constructed by taking three MNIST images, assigning each one of the three colour channels, and stacking the images on top of each other. The resulting images are then padded so that each one of them have size $32 \times 32 \times 3$.

\noindent For Experiment~1, we use $\alpha$ values of 0.5, 5.0, 10.0 and 20.0. For each value of $\alpha$, we train the ($\alpha$, $\alpha$)-GAN and the $(1, \alpha)$-GAN. We additionally train the DCGAN, which corresponds to the $(1, 1)$-GAN. For Experiment 2, we use $k$ values of 0.25, 1.0, 2.0, 7.5 and 15.0. Note that when $k = 2$, we recover LSGAN. For the MNIST dataset, we run 10 trials with the random seeds 123, 500, 1600, 199621, 60677, 20435, 15859, 33764, 79878, and 36123, and train each GAN for 250 epochs. For the RGB datasets (CIFAR-10 and Stacked MNIST), we run 5 trials with the random seeds 123, 1600, 60677, 15859, 79878, and train each GAN for 500 epochs. All experiments utilize an Adam optimzer for the stochastic gradient descent algorithm, with a learning rate of $2 \times 10^{-4}$, and parameters $\beta_ 1 = 0.5$, $\beta_2 = 0.999$ and $\epsilon = 10^{-7}$ \cite{kingma2014adam}.  We also experiment with the addition of a gradient penalty (GP); we add a penalty term to the discriminator's loss function to encourage the discriminator's gradient to have a unit norm \cite{gulrajani2017improved}.

The MNIST experiments were run on one 6130 2.1 GHz 1xV100 GPU, 8 CPUs, and 16 GB of memory. The CIFAR-10 and Stacked MNIST experiments were run on one Epyc 7443 2.8 GHz GPU, 8 CPUs and 16 GB of memory. For each experiment, we report the best overall Fr\'echet Inception Distance (FID) score~\cite{Heusel}, the best average FID score amongst all trials and its variance, and the average epoch the best FID score occurs and its variance. The FID score for each epoch was computed over 10 000 images. For each metric, the lowest numerical value corresponds to the model with the best metric (indicated in bold in the tables). We also report how many trials we include in our summary statistics, as it is possible for a trial to collapse and not train for the full number of epochs. The neural network architectures used in our experiments are presented in Appendix~\ref{sec:architectures}. The training algorithms are presented in Appendix~\ref{sec:algorithms}.

\subsection{Experimental Results}
\begin{figure}[htb!]
    \centering
    \begin{subfigure}[t]{0.47\textwidth}
        \centering
         \includeimages{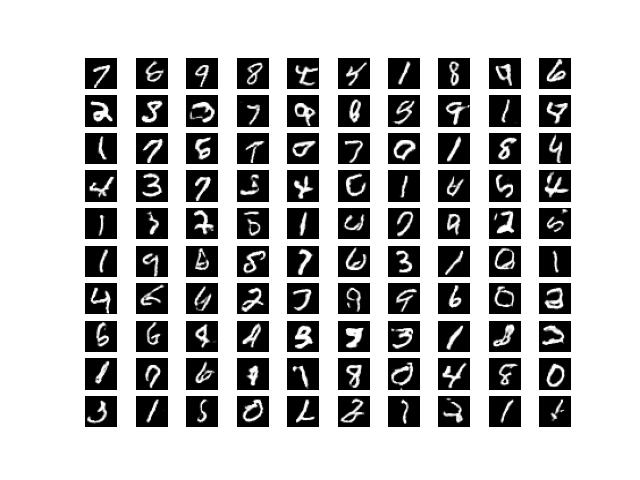}
        \caption{($\alpha_D,\alpha_G$)-GAN for MNIST, $\alpha_D = 1.0$, $\alpha_G = 5.0$, FID: 1.125.}
    \end{subfigure}
\hfill
    \begin{subfigure}[t]{0.47\textwidth}
        \centering
        \includeimages{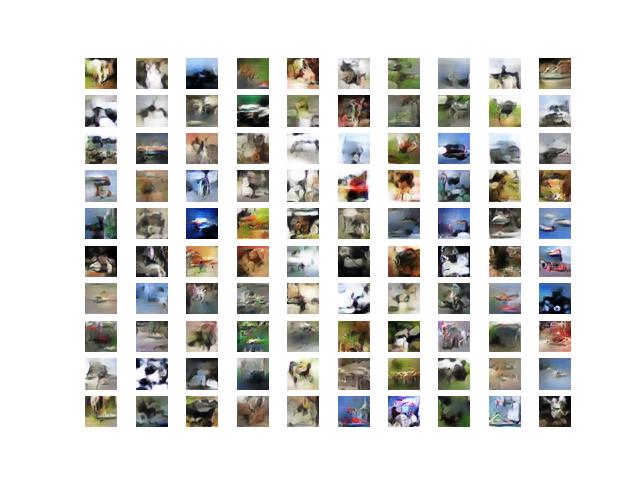}
        \caption{$(\alpha_D,\alpha_G)$-GAN-GP for CIFAR-10, $\alpha_D = 1.0$, $\alpha_G = 20.0$, FID = 8.466.}
    \end{subfigure}
\\ \bigskip
    \centering
    \begin{subfigure}[t]{0.47\textwidth}
        \centering
        \includeimages{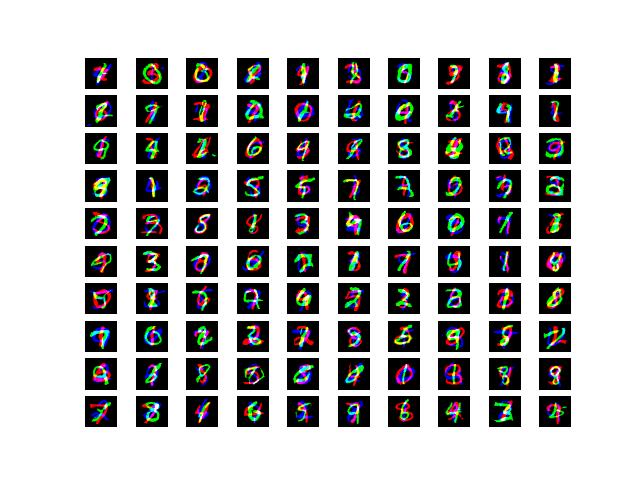}
        \caption{$(\alpha_D,\alpha_G)$-GAN-GP for Stacked MNIST, $\alpha_D = 1.0$, $\alpha_G = 0.5$, FID = 4.833. \\}
    \end{subfigure}
    \caption{Generated images for the best-performing ($\alpha_D$, $\alpha_G$)-GANs.} 
    \label{fig:alphagan_output}
\end{figure}

\begin{figure}[htb!]
    \centering
    \begin{subfigure}[t]{0.47\textwidth}
        \centering
        \includeplot{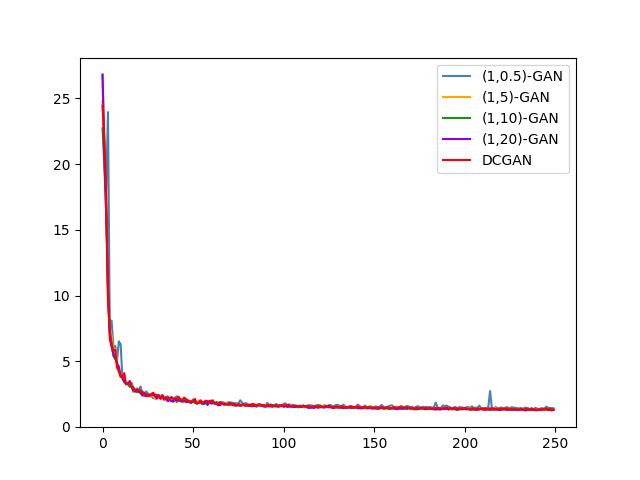}
        \caption{$(1,\alpha)$-GANs for MNIST. \\}
        \label{fig:1alpha_mnist_fid}
    \end{subfigure}
    \hfill
    \begin{subfigure}[t]{0.47\textwidth}
        \centering
        \includeplot{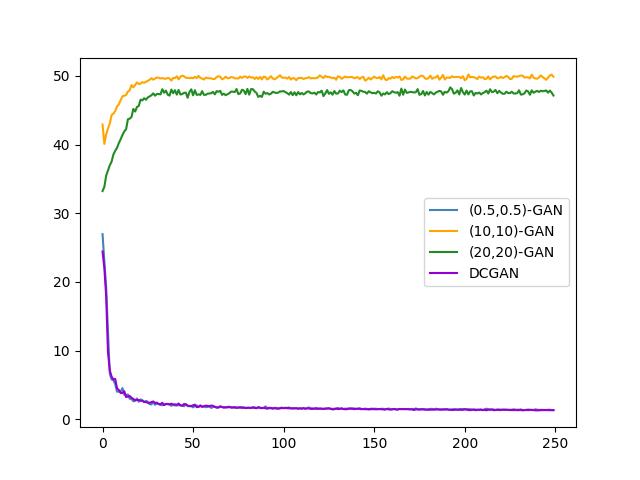}
        \caption{$(\alpha,\alpha)$-GANs for MNIST. \\}
        \label{fig:alpha_mnist_fid}
    \end{subfigure}
    \\
    \begin{subfigure}[t]{0.47\textwidth}
        \includeplot{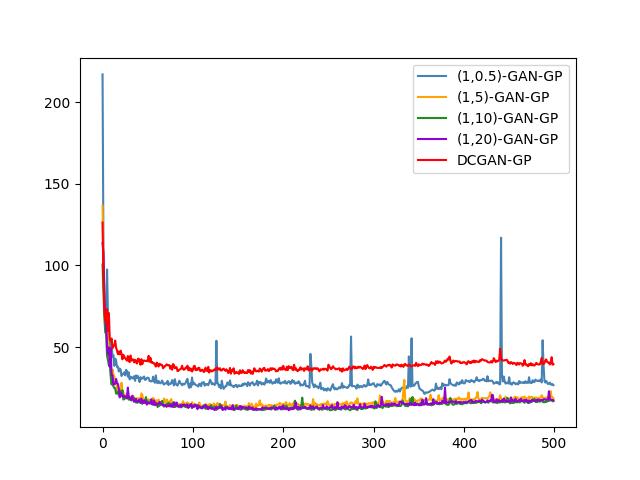}
        \caption{$(1,\alpha)$-GAN-GPs, for CIFAR-10. \\}
        \label{fig:1alpha_cifar_fid}
    \end{subfigure}
    \hfill
    \begin{subfigure}[t]{0.47\textwidth}
        \includeplot{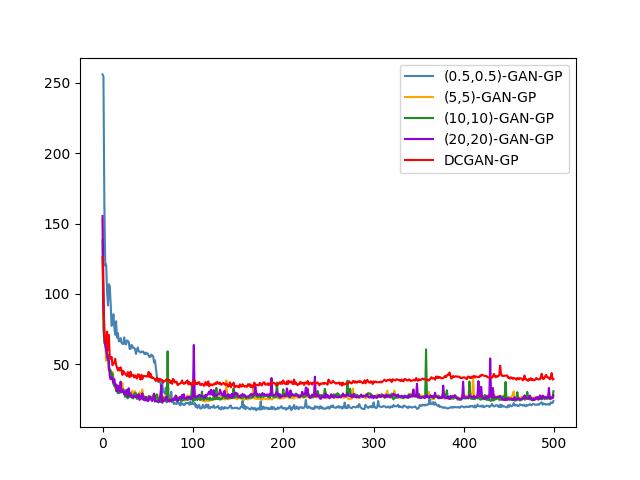}
        \caption{$(\alpha,\alpha)$-GAN-GPs for CIFAR-10. \\}
        \label{fig:alpha_cifar_fid}
    \end{subfigure}
    \\
    \centering
    \begin{subfigure}[t]{0.47\textwidth}
        \includeplot{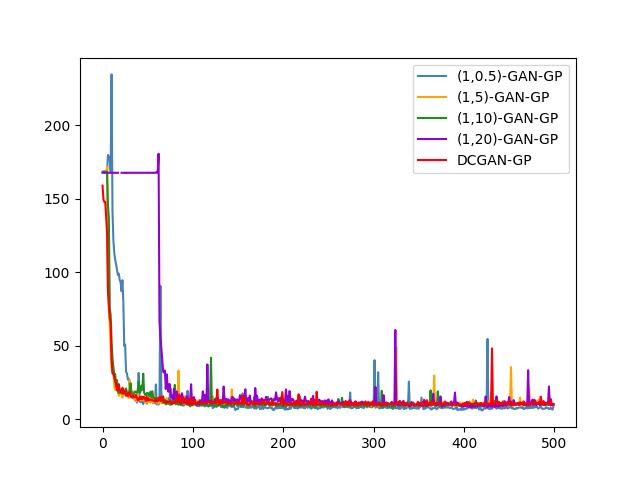}
        \caption{$(1,\alpha)$-GAN-GPs for Stacked MNIST. \\}
        \label{fig:1alpha_smnist_fid}
    \end{subfigure}
    \hfill
    \begin{subfigure}[t]{0.47\textwidth}
        \includeplot{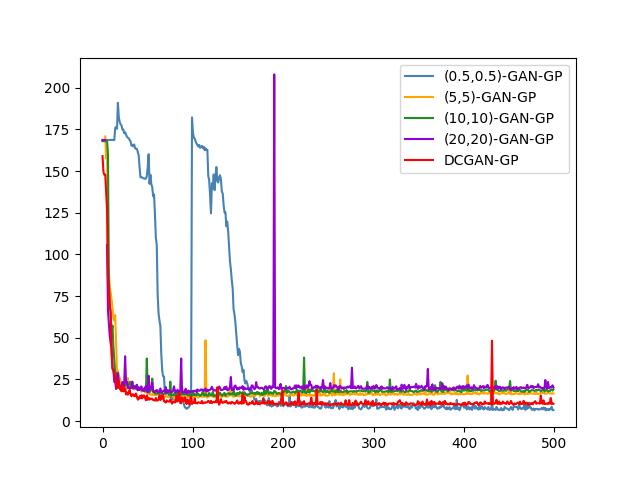}
        \caption{$(\alpha,\alpha)$-GAN-GPs for Stacked MNIST. \\ }
        \label{fig:alpha_smnist_fid}
    \end{subfigure}
    \caption{Average FID scores vs.\ epochs for various $(\alpha_D,\alpha_G)$-GANs.}
    \label{fig:alphagan_fid}
\end{figure}

\begin{figure}
        \centering
        \begin{subfigure}[t]{0.47\textwidth}
            \centering
            \includeimages{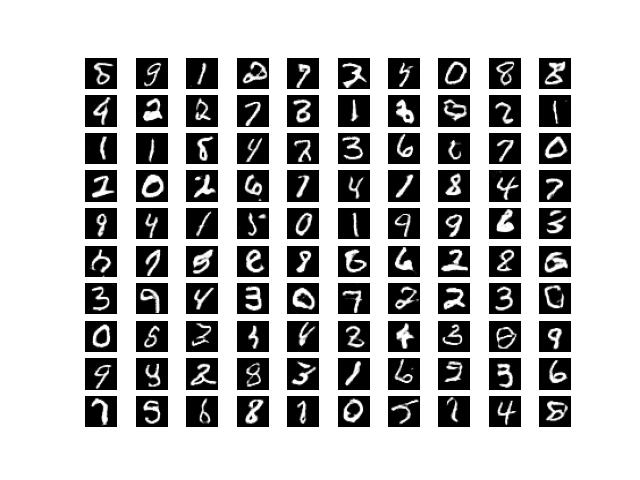}
            \caption{Vanilla-SL$k$GAN-0.25 for MNIST, FID $=1.112$.}
        \end{subfigure}
        \hfill
        \begin{subfigure}[t]{0.47\textwidth}
            \centering
            \includeimages{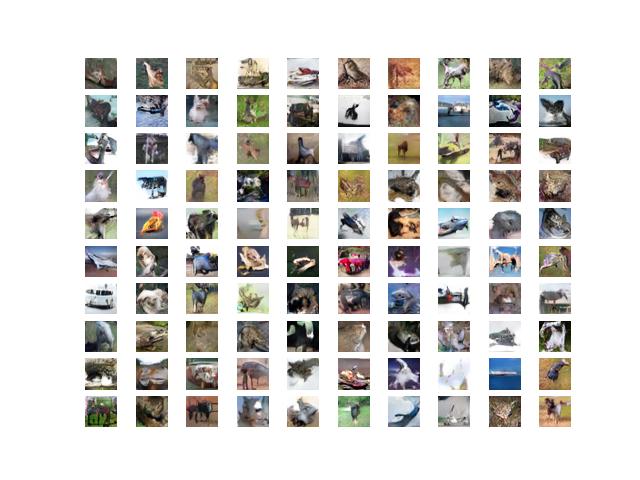}
            \caption{Vanilla-SL$k$GAN-2.0 for CIFAR-10, FID $=4.58$.}
        \end{subfigure}
    \\ \bigskip
        \centering
        \begin{subfigure}[t]{0.47\textwidth}
            \centering
            \includeimages{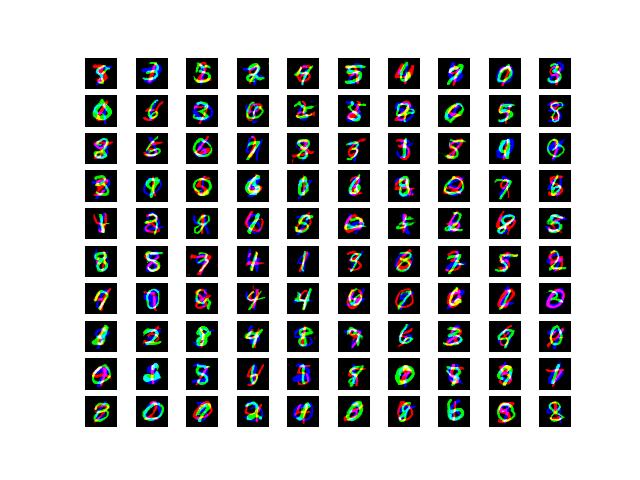}
            \caption{Vanilla-SL$k$GAN-15.0-GP for Stacked MNIST, FID $= 3.836$. \\}
        \end{subfigure}
    \caption{Generated images for best-performing SL$k$GANs.}
    \label{fig:slkgan_output}
    \end{figure}

\begin{figure}[htb]
        \centering
        \begin{subfigure}[t]{0.47\textwidth}
            \includeplot{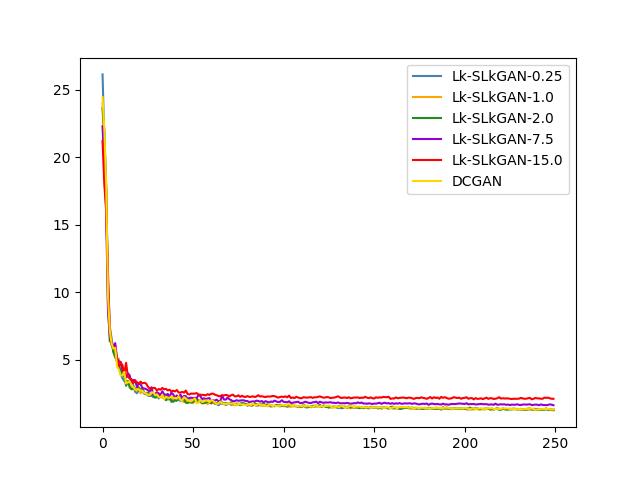}
            \caption{L$k$-SL$k$GANs for MNIST. \\}
            \label{fig:lk_slk_mnist_fid}
        \end{subfigure}
        \hfill
        \begin{subfigure}[t]{0.47\textwidth}
            \includeplot{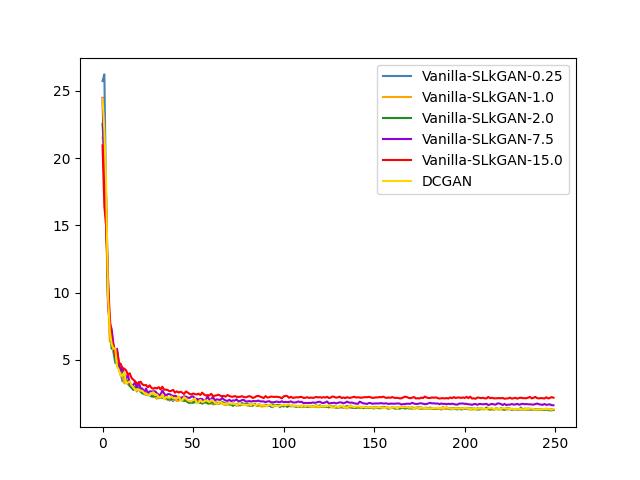}
            \caption{Vanilla-SL$k$GANs for MNIST. \\}
            \label{fig:v_slk_mnist_fid}
        \end{subfigure}
    \\
        \centering
        \begin{subfigure}[t]{0.47\textwidth}
            \includeplot{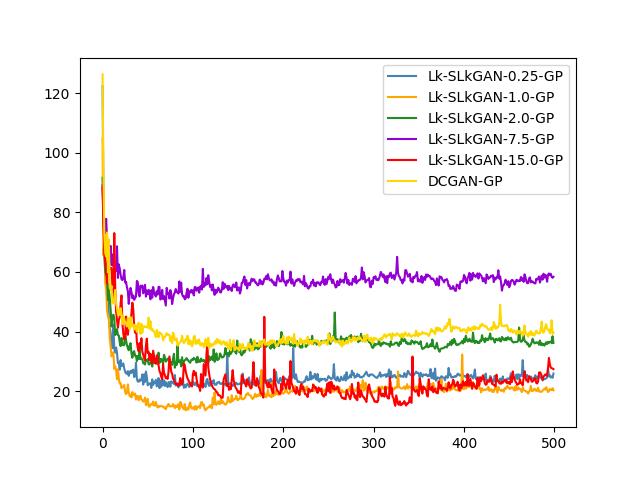}
            \caption{L$k$-SL$k$GAN-GPs for CIFAR-10. \\}
            \label{fig:lk_slk_cifar_fid}
        \end{subfigure}
        \hfill
        \begin{subfigure}[t]{0.47\textwidth}
            \includeplot{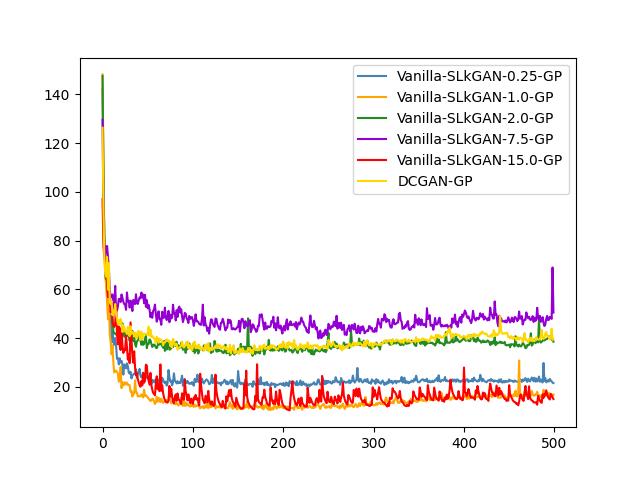}
            \caption{Vanilla-SL$k$GAN-GPs for CIFAR-10. \\}
            \label{fig:v_slk_cifar_fid}
        \end{subfigure}
    \\
        \centering
        \begin{subfigure}[t]{0.47\textwidth}
            \includeplot{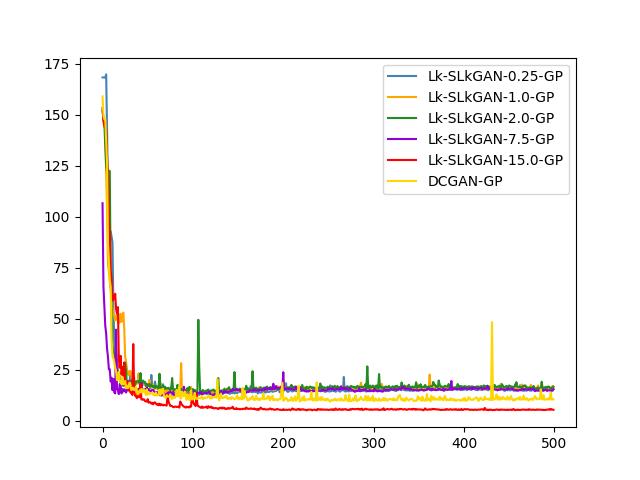}
            \caption{L$k$-SL$k$GAN-GPs for Stacked MNIST. \\}
            \label{fig:lk_slk_smnist_fid}
        \end{subfigure}
        \hfill
        \begin{subfigure}[t]{0.47\textwidth}
            \includeplot{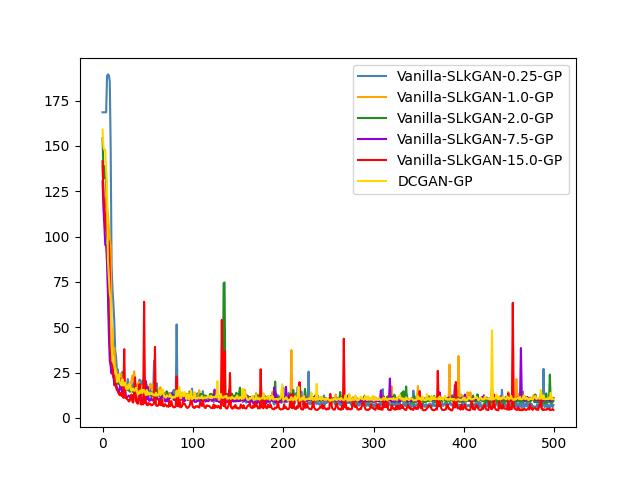}
            \caption{Vanilla-SL$k$GAN-GPs, Stacked MNIST. \\}
            \label{fig:v_slk_smnist_fid}
        \end{subfigure}
        \caption{FID scores vs.\ epochs for various SL$k$GANs.}
        \label{fig:slkgan_fid}
    \end{figure}

We report the FID metrics for Experiment~1 in Tables~\ref{table:alphagan_mnist}, \ref{table:alphagan_cifar10} and~\ref{table:alphagan_smnist}, and for Experiment~2 in Tables~\ref{table:lkgan_mnist}, \ref{table:lkgan_cifar10} and~\ref{table:lkgan_smnist}. We report only on those experiments that produced meaningful results. Models that utilize a simplified gradient penalty have the suffix ``-GP''. We display the output of the best-performing $(\alpha_D, \alpha_G)$-GANs in Figure~\ref{fig:alphagan_output} and the best-performing SL$K$GANs in Figure~\ref{fig:slkgan_output}. Finally, we plot the trajectory of the FID scores throughout training epochs in Figures~\ref{fig:alphagan_fid}~and~\ref{fig:slkgan_fid}.

\bigskip

\begin{table}[htb]
\caption{$(\alpha_D,\alpha_G)$-GAN results for MNIST.} \label{table:alphagan_mnist}
\centering
   \resizebox{\textwidth}{!}{
   \begin{tabular}{c C C N C C C}     
    \toprule
      ($\alpha_D, \alpha_G$)-GAN & Best FID score & Average best FID score & Best FID scores variance & Average epoch & Epoch variance & Number of successful trials (/10) \\ \midrule
    (1,0.5)-GAN & $1.264$ & $1.288$ & $2.979\times 10^{-4}$ & $227.25$ & $420.25$ & 4 \\ 
    (0.5,0.5)-GAN & $1.209$ & $1.265$ & $0.001$ & $234.5$ & $156.7$ & 6 \\ 
    \midrule
    (1,5)-GAN & $\mathbf{1.125}$ & $1.17$ & $8.195\times 10^{-4}$ & $230.3$ & $617.344$ & 10 \\ 
    \midrule
    \textbf{(1,10)-GAN} & $1.147$ & $\mathbf{1.165}$ & $7.984\times 10^{-4}$ & $225.6$ & $253.156$ & 10 \\ 
    (10,10)-GAN & $36.506$ & $39.361$ & $16.312$ & $1.5$ & $0.5$ & 2 \\ 
    \midrule
    (1,20)-GAN & $1.135$ & $1.174$ & $0.001$ & $237.5$ & $274.278$ & 10 \\ 
    (20,20)-GAN & $33.23$ & $33.23$ & $\mathbf{0.0}$ & $\mathbf{1.0}$ & $\mathbf{0.0}$ & 1 \\ 
    \midrule
    DCGAN & $1.154$ & $1.208$ & $0.001$ & $231.3$ & $357.122$ & 10 \\ 
    \bottomrule
    \end{tabular}
  }
    \end{table}

    \begin{table}[h]
    \caption{$(\alpha_D,\alpha_G)$-GAN results for CIFAR-10.}
    \label{table:alphagan_cifar10}
        \centering
        \resizebox{\textwidth}{!}{
       \begin{tabular}{c C C N C C C}
        \toprule
         ($\alpha_D, \alpha_G$)-GAN & Best FID score & Average best FID score & Best FID scores variance & Average epoch & Epoch variance & Number of successful trials (/5) \\ \midrule
        (1,0.5)-GAN-GP & $10.551$ & $14.938$ & $12.272$ & $326.2$ & $1808.7$ & 5 \\ 
        (0.5,0.5)-GAN-GP & $13.734$ & $14.93$ & $0.517$ & $223.6$ & $11378.3$ & 5 \\ 
        \midrule
        (1,5)-GAN-GP & $10.772$ & $11.635$ & $0.381$ & $132.0$ & $1233.5$ & 5 \\ 
        (5,5)-GAN-GP & $20.79$ & $21.72$ & $0.771$ & $\mathbf{84.8}$ & $1527.2$ & 5 \\ 
        \midrule
        \textbf{(1,10)-GAN-GP} & $9.465$ & $\mathbf{10.187}$ & $\mathbf{0.199}$ & $182.6$ & $\mathbf{1096.3}$ & 5 \\ 
        (10,10)-GAN-GP & $19.99$ & $21.095$ & $0.434$ & $131.8$ & $13374.7$ & 5 \\ 
        \midrule
        (1,20)-GAN-GP & $\mathbf{8.466}$ & $10.217$ & $1.479$ & $216.2$ & $6479.7$ & 5 \\ 
        (20,20)-GAN-GP & $19.378$ & $21.216$ & $2.315$ & $138.2$ & $29824.2$ & 5 \\ 
        \midrule
        DCGAN-GP & $25.731$ & $28.378$ & $3.398$ & $158.0$ & $2510.5$ & 5 \\ 
        \bottomrule
        \end{tabular}}
        \end{table}

    \begin{table}[h]
       \caption{$(\alpha_D,\alpha_G)$-GAN results for Stacked MNIST.}
       \label{table:alphagan_smnist}
            \centering
            \resizebox{\textwidth}{!}{
            \begin{tabular}{c C C N C C C}
            \toprule
            ($\alpha_D, \alpha_G$)-GAN & Best FID score & Average best FID score & Best FID scores variance & Average epoch & Epoch variance & Number of successful trials (/5) \\ \midrule
            \textbf{(1,0.5)-GAN-GP} & $\mathbf{4.833}$ & $\mathbf{4.997}$ & $0.054$ & $311.5$ & $23112.5$ & 2 \\ 
            (0.5,0.5)-GAN-GP & $6.418$ & $6.418$ & $\mathbf{0.0}$ & $479.0$ & $\mathbf{0.0}$ & 1 \\ 
            \midrule
            (1,5)-GAN-GP & $7.98$ & $7.988$ & $1.357\times 10^{-4}$ & $379.5$ & $11704.5$ & 2 \\ 
            (5,5)-GAN-GP & $12.236$ & $12.836$ & $0.301$ & $\mathbf{91.5}$ & $387.0$ & 4 \\ 
            \midrule
            (1,10)-GAN-GP & $7.502$ & $7.528$ & $0.001$ & $326.5$ & $14280.5$ & 2 \\ 
            (10,10)-GAN-GP & $14.22$ & $14.573$ & $0.249$ & $95.0$ & $450.0$ & 2 \\ 
            \midrule
            (1,20)-GAN-GP & $8.379$ & $8.379$ & $\mathbf{0.0}$ & $427.0$ & $\mathbf{0.0}$ & 1 \\ 
            (20,20)-GAN-GP & $16.584$ & $16.584$ & $\mathbf{0.0}$ & $94.0$ & $\mathbf{0.0}$ & 1 \\
            \midrule 
            DCGAN-GP & $7.507$ & $7.774$ & $0.064$ & $303.4$ & $11870.8$ & 5 \\ 
            \bottomrule
            \end{tabular}}
    \end{table}

\begin{table}[htb]
\caption{SL$k$GAN results for MNIST.}
\label{table:lkgan_mnist}
    \centering
    \resizebox{\textwidth}{!}{
    \begin{tabular}{c C C N C C C}
    \toprule
    Variant-SL$k$GAN-$k$& Best FID score & Average best FID score & Best FID scores variance & Average epoch & Epoch variance & Number of successful trials (/10) \\ \midrule
    L$k$-SL$k$GAN-0.25 & $1.15$ & $1.174$ & $6.298\times 10^{-4}$ & $224.3$ & $940.9$ & 10 \\ 
    \textbf{Vanilla-SL$\bm{k}$GAN-0.25} & $\mathbf{1.112}$ & $\mathbf{1.162}$ & $0.001$ & $237.0$ & $\mathbf{124.0}$ & 10 \\ 
    \midrule
    L$k$-SL$k$GAN-1.0 & $1.122$ & $1.167$ & $8.857\times 10^{-4}$ & $233.0$ & $124.0$ & 10 \\ 
    Vanilla-SL$k$GAN-1.0 & $1.126$ & $1.17$ & $9.218\times 10^{-4}$ & $226.2$ & $1182.844$ & 10 \\ 
    \midrule
    L$k$-SL$k$GAN-2.0 & $1.148$ & $1.198$ & $5.248\times 10^{-4}$ & $237.2$ & $288.4$ & 10 \\ 
    Vanilla-SL$k$GAN-2.0 & $1.124$ & $1.184$ & $8.933\times 10^{-4}$ & $237.8$ & $138.4$ & 10 \\ 
    \midrule
    L$k$-SL$k$GAN-7.5 & $1.455$ & $1.498$ & $\mathbf{4.422\times 10^{-4}}$ & $229.0$ & $322.222$ & 10 \\ 
    Vanilla-SL$k$GAN-7.5 & $1.439$ & $1.511$ & $0.001$ & $212.2$ & $1995.067$ & 10 \\ 
    \midrule
    L$k$-SL$k$GAN-15.0  & $1.733$ & $1.872$ & $0.005$ & $198.8$ & $1885.733$ & 10 \\ 
    Vanilla-SL$k$GAN-15.0 & $1.773$ & $1.876$ & $0.005$ & $\mathbf{171.6}$ & $3122.267$ & 10 \\ 
    \midrule
    DCGAN & $1.154$ & $1.208$ & $0.001$ & $231.3$ & $357.122$ & 10 \\ 
    \bottomrule
    \end{tabular}}
    \end{table}
    
\begin{table}[hb]
\caption{SL$k$GAN results for CIFAR-10.}
\label{table:lkgan_cifar10}
    \centering
    \resizebox{\textwidth}{!}{
    \begin{tabular}{c C C N C C C}
    \toprule
    Variant-SL$k$GAN-$k$& Best FID score & Average best FID score & Best FID scores variance & Average epoch & Epoch variance & Number of successful trials (/5) \\ \midrule
    L$k$-SL$k$GAN-1.0 & $4.727$ & $118.242$ & $10914.643$ & $\mathbf{60.8}$ & $1897.2$ & 5 \\ 
    Vanilla-SL$k$GAN-1.0 & $4.821$ & $5.159$ & $\mathbf{0.092}$ & $88.0$ & $506.5$ & 5 \\ 
    \midrule
    L$k$-SL$k$GAN-2.0 & $4.723$ & $145.565$ & $7492.26$ & $73.2$ & $3904.2$ & 5 \\ 
    \textbf{Vanilla-SL$\bm{k}$GAN-2.0} & $\mathbf{4.58}$ & $\mathbf{5.1}$ & $0.261$ & $105.4$ & $740.8$ & 5 \\ 
    \midrule
    L$k$-SL$k$GAN-7.5 & $6.556$ & $155.497$ & $7116.521$ & $254.6$ & $18605.3$ & 5 \\ 
    Vanilla-SL$k$GAN-7.5 & $6.384$ & $48.905$ & $8698.195$ & $72.2$ & $1711.7$ & 5 \\ 
    \midrule
    L$k$-SL$k$GAN-15.0 & $8.576$ & $145.774$ & $5945.097$ & $263.0$ & $36463.0$ & 5 \\ 
    Vanilla-SL$k$GAN-15.0 & $7.431$ & $50.868$ & $8753.002$ & $82.6$ & $3106.8$ & 5 \\ 
    \midrule
    DCGAN & $4.753$ & $5.194$ & $0.117$ & $88.6$ & $\mathbf{462.8}$ & 5 \\ 
    \bottomrule
    \rule{0pt}{3ex}L$k$-SL$k$GAN-0.25-GP & $17.366$ & $18.974$ & $2.627$ & $87.8$ & $1897.2$ & 5 \\ 
    Vanilla-SL$k$GAN-0.25-GP & $16.013$ & $17.912$ & $1.961$ & $189.0$ & $9487.5$ & 5 \\
    \midrule 
    L$k$-SL$k$GAN-1.0-GP & $10.771$ & $12.567$ & $1.083$ & $77.8$ & $\mathbf{239.2}$ & 5 \\ 
    Vanilla-SL$k$GAN-1.0-GP & $8.569$ & $9.588$ & $\mathbf{0.749}$ & $197.6$ & $2690.3$ & 5 \\ 
    \midrule
    L$k$-SL$k$GAN-2.0-GP & $23.11$ & $25.013$ & $1.924$ & $\mathbf{75.4}$ & $658.8$ & 5 \\ 
    Vanilla-SL$k$GAN-2.0-GP & $28.215$ & $29.69$ & $1.242$ & $232.0$ & $20438.5$ & 5 \\ 
    \midrule
    L$k$-SL$k$GAN-7.5-GP & $33.304$ & $41.48$ & $49.187$ & $82.8$ & $1081.2$ & 5 \\ 
    Vanilla-SL$k$GAN-7.5-GP & $33.085$ & $34.799$ & $1.597$ & $290.8$ & $12714.7$ & 5 \\ 
    \midrule
    L$k$-SL$k$GAN-15.0-GP & $9.157$ & $12.504$ & $3.839$ & $310.4$ & $6976.8$ & 5 \\ 
    \textbf{Vanilla-SL$\bm{k}$GAN-15.0-GP} & $\mathbf{7.283}$ & $\mathbf{8.568}$ & $1.535$ & $185.6$ & $5978.3$ & 5 \\ 
    \midrule
    DCGAN-GP & $25.731$ & $28.378$ & $3.398$ & $158.0$ & $2510.5$ & 5 \\ 
    \bottomrule
   \end{tabular}}
    \end{table}
    
\begin{table}[htb]
\caption{SL$k$GAN results for Stacked MNIST.}
\label{table:lkgan_smnist}
    \centering
    \resizebox{\textwidth}{!}{
    \begin{tabular}{c C C N C C C}
    \toprule
    Variant-SL$k$GAN-$k$ & Best FID score & Average best FID score & Best FID scores variance & Average epoch & Epoch variance & Number of successful trials (/5) \\ \midrule
    L$k$-SL$k$GAN-0.25-GP & $10.541$ & $11.824$ & $0.678$ & $113.6$ & $356.3$ & 5 \\ 
    Vanilla-SL$k$GAN-0.25-GP & $5.197$ & $5.197$ & $\mathbf{0.0}$ & $496.0$ & $\mathbf{0.0}$ & 1 \\ 
    \midrule
    L$k$-SL$k$GAN-1.0-GP & $11.545$ & $12.046$ & $0.291$ & $\mathbf{89.0}$ & $238.5$ & 5 \\ 
    Vanilla-SL$k$GAN-1.0-GP & $7.475$ & $7.626$ & $0.045$ & $177.0$ & $3528.0$ & 2 \\ 
    \midrule
    L$k$-SL$k$GAN-2.0-GP & $10.682$ & $12.782$ & $2.12$ & $180.2$ & $28484.7$ & 5 \\ 
    Vanilla-SL$k$GAN-2.0-GP & $6.023$ & $7.096$ & $0.991$ & $416.667$ & $12244.333$ & 3 \\ 
    \midrule
    L$k$-SL$k$GAN-7.5-GP & $8.912$ & $9.906$ & $0.577$ & $239.0$ & $35663.5$ & 5 \\ 
    Vanilla-SL$k$GAN-7.5-GP & $6.074$ & $6.43$ & $0.164$ & $238.0$ & $21729.5$ & 5 \\ 
    \midrule
    L$k$-SL$k$GAN-15.0-GP & $4.458$ & $4.74$ & $0.029$ & $253.4$ & $11512.3$ & 5 \\ 
    \textbf{Vanilla-SL$\bm{k}$GAN-15.0-GP} & $\mathbf{3.836}$ & $\mathbf{3.873}$ & $0.002$ & $485.0$ & $354.667$ & 4 \\ 
    \midrule
    DCGAN-GP & $7.507$ & $7.774$ & $0.064$ & $303.4$ & $11870.8$ & 5 \\ 
    \bottomrule
    \end{tabular}}
    \end{table}
    
    \newpage

\vfill
\clearpage

    \vfill
    \clearpage

    \subsection{Discussion}
    \subsubsection{Experiment~1}
    From Table~\ref{table:alphagan_mnist}, we note that 37 of the 90 trials collapse before 250 epochs have passed without a gradient penalty. The (5,5)-GAN collapses for all 5 trials, and hence it is not displayed in Table~\ref{table:alphagan_mnist}. This behaviour is expected, as the ($\alpha$,$\alpha$)-GAN is more sensitive to exploding gradients when $\alpha$ does not tend to 0 or $+\infty$ \cite{sankar2021}. The addition of a gradient penalty could mitigate the discriminator's gradients diverging in the (5,5)-GAN by encouraging gradients to have a unit norm. Using a VanillaGAN discriminator  with an $\alpha$-GAN generator (i.e., the (1,$\alpha$)-GAN) produces better quality images for all tested values of $\alpha$, compared to when both networks utilize an $\alpha$-GAN loss function. The (1,10)-GAN achieves excellent stability, converging in all 10 trials, and also achieves the lowest average FID score. The (1,5)-GAN achieves the lowest FID score overall, marginally outperforming DCGAN. Note that when the average best FID score is very close to the best FID score, the resulting best FID score variance is quite small (of the order of $10^{-3}$), indicating little statistical variability over the trials.
    
    Likewise, for the CIFAR-10 and Stacked MNIST datasets, the (1,$\alpha$)-GAN produces lower FID scores than the $(\alpha, \alpha)$-GAN (see Tables~\ref{table:alphagan_cifar10} and~\ref{table:alphagan_smnist}). However, both models are more stable with the CIFAR-10 dataset. With the exception of DCGAN, no model converged to its best FID score for all 5 trials with the Stacked MNIST dataset. Comparing the trials that did converge, both $(\alpha,\alpha)$-GAN and $(1,\alpha)$-GAN performed better on the Stacked MNIST dataset than the CIFAR-10 dataset. For CIFAR-10, the (1,10)- and (1,20)-GANs produced the best overall FID score and the best average FID score respectively. On the other hand, the (1,0.5)-GAN produced the best overall FID score and the best average FID score for the Stacked MNIST dataset. We also observe a tradeoff between speed and performance for the CIFAR-10 and Stacked MNIST datasets: the $(1,\alpha)$-GANs arrive at their lowest FID scores later than their respective $(\alpha, \alpha)$-GANs, but achieve lower FID scores overall.
    
   Comparing Figures~\ref{fig:1alpha_cifar_fid} and~\ref{fig:alpha_cifar_fid}, we observe that the $(\alpha, \alpha)$-GAN-GP provides more stability than the $(1,\alpha)$-GAN for lower values of $\alpha$ (i.e. $\alpha = 0.5)$, while the $(1, \alpha)$-GAN-GP exhibits more stability for higher $\alpha$ values ($\alpha = 10$ and $\alpha = 20$). Figures \ref{fig:1alpha_smnist_fid} and \ref{fig:alpha_smnist_fid} show that the two $\alpha$-GANs trained on the Stacked MNIST dataset exhibit unstable behaviour earlier into training when $\alpha = 0.5$ or $\alpha = 20$. However, both systems stabilize and converge to their lowest FID scores as training progresses. The (0.5,0.5)-GAN-GP system in particular exhibits wildly erratic behaviour for the first 200 epochs, then finishes training with a stable trajectory that outperforms DCGAN-GP.
    
    A future direction is to explore how the complexity of an image dataset influences the best choice of $\alpha$. For example, the Stacked MNIST dataset might be considered to be less complex than CIFAR-10, as images in the Stacked MNIST dataset only contain four unique colours (black, red, green, and blue), while the CIFAR-10 dataset utilizes significantly more colours.
    
    \subsubsection{Experiment~2}
    We see from Table~\ref{table:lkgan_mnist} that all L$k$-L$k$GANs and Vanilla-SL$k$GANs have FID scores comparable to the DCGAN. When $k = 15$, Vanilla-SL$k$GAN and L$k$-SL$k$GAN arrive at their lowest FID scores slightly earlier than DCGAN and other SL$k$GANs.
    
    The addition of a simplified gradient penalty is necessary for L$k$-SL$k$GAN to achieve overall good performance on the CIFAR-10 dataset (see Table~\ref{table:lkgan_cifar10}). Interestingly, Vanilla-SL$k$GAN achieves lower FID scores without a gradient penalty for lower $k$ values ($k = 1, 2$), and with a gradient penalty for higher $k$ values ($k = 7.5, 15$). When $k = 0.25$, both SL$k$GANs collapsed for all 5 trials without a gradient penalty.
    
    Table~\ref{table:lkgan_smnist} shows that Vanilla-SL$k$GANs achieve better FID scores than their respective L$k$-L$k$GAN counterparts. However, L$k$-L$k$GANs are more stable, as no single trial collapsed, while 10 of the 25 Vanilla-SL$k$GAN trials collapsed before 500 epochs had passed. While all Vanilla-SL$k$GANs outperform the DCGAN with gradient penalty, L$k$-SL$k$GAN-GP only outperforms DCGAN-GP when $k = 15$. Except for when $k = 7.5$, we observe that the L$k$-SL$k$GAN system takes less epochs to arrive at its lowest FID score. Comparing Figures~\ref{fig:lk_slk_smnist_fid} and~\ref{fig:v_slk_smnist_fid}, we observe that L$k$-SL$k$GANs exhibit more stable FID score trajectories than their respective Vanilla-SL$k$GANs. This makes sense, as the L$k$GAN loss function aims to increase the GAN's stability compared to DCGAN~\cite{Bhatia_2021}.

\section{Conclusion}\label{sec:conclusion}
    We introduced a parameterized CPE-based generator loss function for a dual-objective GAN termed $\La$-GAN which, when used in tandem with a canonical discriminator loss function that achieves its optimum in~\eqref{eq:optimal_discriminator}, minimizes a Jensen-$f_\alpha$-divergence. We showed that this system can recover VanillaGAN, $(1,\alpha)$-GAN, and L$k$GAN as special cases. We conducted experiments with the three aforementioned $\La$-GANs on three image datasets. The experiments indicate that $(1,\alpha)$-GAN exhibits better performance than $(\alpha,\alpha)$-GAN with $\alpha>1$. They also show that the devised SL$k$GAN system achieves lower FID scores with a VanillaGAN discriminator compared with an L$k$GAN discriminator. 
    
    Future work consists of unveiling more examples of existing GANs that fall under our result as well as applying $\La$-GAN to novel judiciously designed CPE losses $\La$ and evaluating the performance (in terms of both quality and diversity of generated samples) and the computational efficiency of the resulting models. Another interesting and related direction is to study $\La$-GAN within the context of $f$-GANs, given that the Jensen-$f$-divergence is itself an $f$-divergence (see Remark~\ref{jensen-f-div-is-an-f-div}), by systematically analyzing different Jensen-$f$-divergences and the role they play in improving GAN performance and stability. Other worthwhile directions include incorporating the proposed $\La$~loss into state-of-the-art GAN models, such as among others  BigGAN~\cite{brock2018large}, StyleGAN~\cite{karras2019style} and CycleGAN~\cite{almahairi2018augmented}, for high-resolution data generation and image-to-image translation applications, conducting a meticulous analysis of the sensitivity of the models' performance to different values of the $\alpha$ parameter and providing guidelines on how best to tune $\alpha$ for different types of datasets.

\bigskip\noindent
\section*{Acknowledgment}
This work was supported in part by the Natural Sciences and Engineering Research Council (NSERC) of Canada.

\vspace{10pt}


\appendix
\section{Neural Network Architectures}\label{sec:architectures}
We outline the architectures used for the generator and discriminator. For the MNIST dataset, we use the architectures of~\cite{Bhatia_2021}. For the CIFAR-10 and Stacked MNIST datasets, we base the architectures on~\cite{Radford2015}. We summarize some aliases for the architectures in Table~\ref{table:arch_alias}. For all models we use a batch size of 100 and noise size of 784 for the generator input.

\bigskip 

\begin{table}[htb]
\caption{Summary of aliases used to describe neural network architectures.}
\label{table:arch_alias} 
    \centering
    \begin{tabular}{c c}
        \toprule
        Alias & Definition\\
        \midrule
        FC & Fully Connected\\
        UpConv2D & Deconvolutional Layer\\
        Conv2D & Convolutional Layer\\
        BN & Batch Normalization\\
        LeakyReLU & Leaky Rectified Linear Unit\\
        \bottomrule
    \end{tabular}
\end{table}

We omit the bias in the convolutional and deconvolutional layers to decrease the number of parameters being trained, which in turn decreases computation times. We initialize our kernels using a normal distribution with zero mean and variance 0.01. We present the MNIST architectures in Tables~\ref{table:mnist_arch_dis} and~\ref{table:mnist_arch_gen}, and the CIFAR-10 and Stacked MNIST architectures in Tables~\ref{table:rgb_arch_dis} and~\ref{table:rgb_arch_gen}. 

\bigskip



\begin{table}[htb!]
    \centering
    \caption{Discriminator architecture for the MNIST dataset.} 
    \label{table:mnist_arch_dis}
    \begin{tabular}{c c c c c c}
        \toprule
        Layer & Output Size & Kernel & Stride & BN & Activation\\
        \midrule
        Input & $28 \times 28 \times 1$ &No & & \\
        Conv2D & $14 \times 14 \times 64$ & $5 \times 5$ & 2 & No & LeakyReLU(0.3)\\
        Dropout(0.3)& & & &No &\\
        Conv2D & $7 \times 7 \times 128$ & $5 \times 5$ & 2 & No & LeakyReLU(0.3)\\
        Dropout(0.3)& & & &No &\\
        FC & 1 & & &No & Sigmoid\\
        \bottomrule
    \end{tabular}
\end{table}

\begin{table}[htb!]
    \centering
    \caption{Generator architecture for the MNIST dataset.}
    \label{table:mnist_arch_gen}
    \begin{tabular}{c c c c c c}
        \toprule
        Layer & Output Size & Kernel & Stride & BN & Activation\\
        \midrule
        Input & $784$ & & & \\
        FC & $7 \times 7 \times 256$ & & &\\
        UpConv2D & $7 \times 7 \times 128$ & $5 \times 5$ & 1 & Yes & LeakyReLU(0.3)\\
        UpConv2D & $14 \times 14 \times 64$ & $5 \times 5$ & 2 & Yes & LeakyReLU(0.3)\\
        UpConv2D & $28 \times 28 \times 1$ & $5 \times 5$ & 2 & No & Tanh\\
        \bottomrule
    \end{tabular}
\end{table}

\begin{table}[ht]
\caption{Discriminator architecture for the CIFAR-10 and Stacked MNIST datasets.}
    \label{table:rgb_arch_dis}
    \centering
    \begin{tabular}{c c c c c c}
        \toprule
        Layer & Output Size & Kernel & Stride & BN & Activation\\
        \midrule
        Input & $32 \times 32 \times 3$ & & & \\
        Conv2D & $16 \times 16 \times 128$ & $3 \times 3$ & 2 & No & LeakyReLU(0.2)\\
        Conv2D & $8 \times 8 \times 128$ & $3 \times 3$ & 2 & No & LeakyReLU(0.2)\\
        Conv2D & $4 \times 4 \times 256$ & $3 \times 3$ & 2 & No & LeakyReLU(0.2)\\
        Dropout(0.4)& & & &No &\\
        FC & 1 & & &  & Sigmoid\\
        \bottomrule
    \end{tabular}
    \end{table}

\begin{table}[hb!]
\caption{Generator architecture for the CIFAR-10 and Stacked MNIST datasets.}
    \label{table:rgb_arch_gen}
    \centering
    \begin{tabular}{c c c c c c}
        \toprule
        Layer & Output Size & Kernel & Stride & BN & Activation\\
        \midrule
        Input & $784$ & & & \\
        FC & $4 \times 4 \times 256$ & & &\\
        UpConv2D & $8 \times 8 \times 128$ & $4 \times 4$ & 2 & Yes & LeakyReLU(0.2)\\
        UpConv2D & $16 \times 16 \times 128$ & $4 \times 4$ & 2 & Yes & LeakyReLU(0.2)\\
        UpConv2D & $32 \times 32 \times 128$ & $4 \times 4$ & 2 & Yes & LeakyReLU(0.2)\\
        Conv2D & $32 \times 32 \times 3$ & $3 \times 3$ & 1 & No & Tanh\\
        \bottomrule
    \end{tabular}
\end{table}



\clearpage

\section{Algorithms}\label{sec:algorithms}
We outline the algorithms used to train our models in Algorithms~\ref{algo:alphagan}, \ref{algo:lk_slkgan} and~\ref{algo:vanilla_slkgan}.

\bigskip

\begin{algorithm}
    \caption{Overview of ($\alpha_D$, $\alpha_G$)-GAN training}
    \label{algo:alphagan}
    \begin{algorithmic}[htb]
        \State {\bfseries{Require}} $\alpha_D$, $\alpha_G$, Number of epochs $n_e$, Batch size $B$, Learning rate $\eta$
        \State {\bfseries{Initialize}} Generator $G$ with parameters $\btheta_G$, Discriminator $D$ with parameters $\btheta_D$.
        \For{$i = 1$ to $n_e$}
          
            \State {\bfseries{Sample}} batch of real data $\x = \{\x_1,..., \x_B\}$ from dataset
            \State {\bfseries{Sample}} batch of Gaussian noise vectors $\z = \{\z_1,...,\z_B\} \sim \mathcal{N}(\textbf{0}, \textbf{I})$
            \State {\bfseries{Update}} the discriminator's parameters using an Adam optimizer with learning rate $\eta$ by descending the gradient:
                \begin{align*}
                    \nabla_{{\btheta}_D}\left(-\frac{1}{B}\sum_{i=1}^{B}(-\ell_\alpha(1, D(\x_i)) - \ell_\alpha(0, D(G(\z_i))))\right)
                \end{align*}
            \State or {\bf update} the discriminator's parameters with a simplified GP:
                \begin{align*}
                    &\nabla_{{\btheta}_D}\left(-\frac{1}{B}\sum_{i=1}^{B}(-\ell_\alpha(1, D(\x_i)) - \ell_\alpha(0, D(G(\z_i))))\right.\\
                    &\quad\left.+5\left(\sum_{i=1}^{B}\bigg{|}\bigg{|}\nabla_{\x}\log\left(\frac{D(\x)}{1-D(\x)}\right)\bigg{|}\bigg{|}_{2}^{2}\right)\right)
                \end{align*}
            \State {\bfseries{Update}} the generator's parameters using an Adam optimizer with learning rate $\eta$ and descending the gradient:
                \begin{align*}
                    &\nabla_{{\btheta}_G}\left(\frac{1}{B}\sum_{i=1}^{B} \ell_\alpha(0, D(G(\z_i)))\right)
                \end{align*}   
        \EndFor
    \end{algorithmic}
\end{algorithm}


\vspace{6pt} 

\begin{algorithm}
    \caption{Overview of L$k$-SL$k$GAN training}
    \label{algo:lk_slkgan}
    \begin{algorithmic}[hb]
        \State {\bfseries{Require}} $k$, Number of epochs $n_e$, Batch size $B$, Learning rate $\eta$
        \State {\bfseries{Initialize}} Generator $G$ with parameters $\btheta_G$, Discriminator $D$ with parameters $\btheta_D$.
        \For{$i = 1$ to $n_e$}
          
            \State {\bfseries{Sample}} batch of real data $\x = \{\x_1,..., \x_B\}$ from dataset
            \State {\bfseries{Sample}} batch of Gaussian noise vectors $\z = \{\z_1,...,\z_B\} \sim \mathcal{N}(\textbf{0}, \textbf{I})$
            \State {\bfseries{Update}} the discriminator's parameters using an Adam optimizer with learning rate $\eta$ by descending the gradient:
                \begin{align*}
                    \nabla_{{\btheta}_D}\left(\frac{1}{B}\sum_{i=1}^{B}\left(\frac{1}{2}(D(\x_i) - 1)^2 + \frac{1}{2}(D(G(\z_i))^2)\right)\right)
                \end{align*}
            \State or {\bf update} the discriminator's parameters with a simplified GP:
                \begin{align*}
                    &\nabla_{{\btheta}_D}\left(\frac{1}{B}\sum_{i=1}^{B}\left(\frac{1}{2}(D(\x_i) - 1)^2 + \frac{1}{2}(D(G(\z_i))^2)\right)\right.\\
                    &\quad\left.+5\left(\sum_{i=1}^{B}\bigg{|}\bigg{|}\nabla_{\x}\log\left(\frac{D(\x)}{1-D(\x)}\right)\bigg{|}\bigg{|}_{2}^{2}\right)\right)
                \end{align*}
            \State {\bfseries{Update}} the generator's parameters using an Adam optimizer with learning rate $\eta$ and descending the gradient:
                \begin{align*}
                    &\nabla_{{\btheta}_G}\left(\frac{1}{B}\sum_{i=1}^{B} \frac{1}{2}(|1 - D(G(\z_i))|^k - 1)\right)
                \end{align*}
            
        \EndFor
    \end{algorithmic}
\end{algorithm}
\vfill

\begin{algorithm}
    \caption{Overview of Vanilla-SL$k$GAN training}
    \label{algo:vanilla_slkgan}
    \begin{algorithmic}[htb]
        \State {\bfseries{Require}} $k$, Number of epochs $n_e$, Batch size $B$, Learning rate $\eta$
        \State {\bfseries{Initialize}} Generator $G$ with parameters $\btheta_G$, Discriminator $D$ with parameters $\btheta_D$.
        \For{$i = 1$ to $n_e$}
          
            \State {\bfseries{Sample}} batch of real data $\x = \{\x_1,..., \x_B\}$ from dataset
            \State {\bfseries{Sample}} batch of noise vectors $\z = \{\z_1,...,\z_B\} \sim \mathcal{N}(\textbf{0}, \textbf{I})$
            \State {\bfseries{Update}} the discriminator's parameters using an Adam optimizer with learning rate $\eta$ by descending the gradient:
                \begin{align*}
                    \nabla_{{\btheta}_D}\left(-\frac{1}{B}\sum_{i=1}^{B}\left(\log(D(\x_i)) + \log(1 - D(G(\z_i)))\right)\right)
                \end{align*}
            \State or {\bf update} the discriminator's parameters with a simplified (GP):
                \begin{align*}
                    &\nabla_{{\btheta}_D}\left(-\frac{1}{B}\sum_{i=1}^{B}\left(\log(D(\x_i)) + \log(1 - D(G(\z_i)))\right)\right.\\
                    &\quad\left.+5\left(\sum_{i=1}^{B}\bigg{|}\bigg{|}\nabla_{\x}\log\left(\frac{D(\x)}{1-D(\x)}\right)\bigg{|}\bigg{|}_{2}^{2}\right)\right)
                \end{align*}
            \State {\bfseries{Update}} the generator's parameters using an Adam optimizer with learning rate $\eta$ and descending the gradient:
                \begin{align*}
                    &\nabla_{{\btheta}_G}\left(\frac{1}{B}\sum_{i=1}^{B} \frac{1}{2}(|1 - D(G(\z_i))|^k - 1)\right)
                \end{align*}
            
        \EndFor
    \end{algorithmic}
\end{algorithm}


\bibliographystyle{plain}
\bibliography{citations}

\begin{thebibliography}{10}

\bibitem{silvey}
S.~M. Ali and S.~D. Silvey.
\newblock A general class of coefficients of divergence of one distribution from another.
\newblock {\em Journal of the Royal Statistical Society. Series B (Methodological)}, 28(1):131--142, 1966.

\bibitem{almahairi2018augmented}
Amjad Almahairi, Sai Rajeshwar, Alessandro Sordoni, Philip Bachman, and Aaron Courville.
\newblock Augmented {C}ycle{GAN}: Learning many-to-many mappings from unpaired data.
\newblock In {\em Proceedings of the International Conference on Machine Learning}, pages 195--204. PMLR, 2018.

\bibitem{ARIMOTO1971181}
Suguru Arimoto.
\newblock {I}nformation-theoretical considerations on estimation problems.
\newblock {\em Information and Control}, 19(3):181--194, 1971.

\bibitem{arjovsky2017wasserstein}
Martin Arjovsky, Soumith Chintala, and L{\'e}on Bottou.
\newblock {W}asserstein generative adversarial networks.
\newblock In {\em Proceedings of the International Conference on Machine Learning}, pages 214--223. PMLR, 2017.

\bibitem{Bhatia_2021}
Himesh Bhatia, William Paul, Fady Alajaji, Bahman Gharesifard, and Philippe Burlina.
\newblock {L}east $k$th-order and {R}\'enyi generative adversarial networks.
\newblock {\em Neural Computation}, 33(9):2473--2510, 2021.

\bibitem{brock2018large}
Andrew Brock, Jeff Donahue, and Karen Simonyan.
\newblock Large scale {GAN} training for high fidelity natural image synthesis.
\newblock {\em arXiv preprint arXiv:1809.11096}, 2018.

\bibitem{csi}
Imre Csiszar.
\newblock Eine {I}nformationstheoretische {U}ngleichung und ihre {A}nwendung auf den {B}ewis der {E}rgodizitat on {M}arkhoffschen {K}etten.
\newblock {\em Publications of the Mathematical Institute of the Hungarian Academy of Sciences, Series A}, 8, 01 1963.

\bibitem{csiszar67}
Imre Csisz\'{a}r.
\newblock Information-type measures of difference of probability distributions and indirect observations.
\newblock {\em Studia Sci. Math. Hungarica}, 2:299--318, 1967.

\bibitem{mnist}
Li~Deng.
\newblock {T}he {M}{N}{I}{S}{T} database of handwritten digit images for machine learning research.
\newblock {\em IEEE Signal Processing Magazine}, 29(6):141--142, 2012.

\bibitem{goodfellow2020generative}
Ian Goodfellow, Jean Pouget-Abadie, Mehdi Mirza, Bing Xu, David Warde-Farley, Sherjil Ozair, Aaron Courville, and Yoshua Bengio.
\newblock {G}enerative adversarial nets.
\newblock In Z.~Ghahramani, M.~Welling, C.~Cortes, N.~Lawrence, and K.Q. Weinberger, editors, {\em Advances in Neural Information Processing Systems}, volume~27, pages 2672--2680. Curran Associates, Inc., 2014.

\bibitem{gulrajani2017improved}
Ishaan Gulrajani, Faruk Ahmed, Martin Arjovsky, Vincent Dumoulin, and Aaron~C Courville.
\newblock {I}mproved training of {W}asserstein {G}{A}{N}s.
\newblock {\em Advances in Neural Information Processing Systems}, 30, 2017.

\bibitem{Hellinger+1909+210+271}
E.~Hellinger.
\newblock {\em Journal für die reine und angewandte Mathematik}, 1909(136):210--271, 1909.

\bibitem{Heusel}
Martin Heusel, Hubert Ramsauer, Thomas Unterthiner, Bernhard Nessler, and Sepp Hochreiter.
\newblock {{G}{A}{N}}s trained by a two time-scale update rule converge to a local {N}ash equilibrium.
\newblock In {\em Advances in {N}eural {I}nformation {P}rocessing {S}ystems}, pages 6626--6637, 2017.

\bibitem{jordon2018pate}
James Jordon, Jinsung Yoon, and Mihaela Van Der~Schaar.
\newblock {P}{A}{T}{E}-{G}{A}{N}: {G}enerating synthetic data with differential privacy guarantees.
\newblock In {\em Proceedings of the International Conference on Learning Representations}, 2018.

\bibitem{karras2019style}
Tero Karras, Samuli Laine, and Timo Aila.
\newblock A style-based generator architecture for generative adversarial networks.
\newblock In {\em Proceedings of the IEEE/CVF Conference on Computer Vision and Pattern Recognition}, pages 4401--4410, 2019.

\bibitem{kingma2014adam}
Diederik Kingma and Jimmy Ba.
\newblock {A}dam: {A} method for stochastic optimization.
\newblock In {\em Proceedings of the International Conference on Learning Representations}, 2014.

\bibitem{cifar10}
Alex Krizhevsky, Geoffrey Hinton, et~al.
\newblock {L}earning multiple layers of features from tiny images.
\newblock 2009.

\bibitem{kl_div}
Solomon Kullback and Richard~A Leibler.
\newblock {O}n information and sufficiency.
\newblock {\em The Annals of Mathematical Statistics}, 22(1):79--86, 1951.

\bibitem{sankar2021}
Gowtham~R Kurri, Tyler Sypherd, and Lalitha Sankar.
\newblock {R}ealizing {G}{A}{N}s via a tunable loss function.
\newblock In {\em Proceedings of the IEEE Information Theory Workshop (ITW)}, pages 1--6, 2021.

\bibitem{sankar2022}
Gowtham~R Kurri, Monica Welfert, Tyler Sypherd, and Lalitha Sankar.
\newblock $\alpha$-{G}{A}{N}: {C}onvergence and estimation guarantees.
\newblock In {\em Proceedings of the IEEE International Symposium on Information Theory (ISIT)}, pages 276--281, 2022.

\bibitem{kwon2019predicting}
Yong-Hoon Kwon and Min-Gyu Park.
\newblock {P}redicting future frames using retrospective cycle {G}{A}{N}.
\newblock In {\em Proceedings of the IEEE/CVF Conference on Computer Vision and Pattern Recognition (CVPR)}, June 2019.

\bibitem{liese2006divergences}
F.~Liese and I.~Vajda.
\newblock {O}n divergences and informations in statistics and information theory.
\newblock {\em IEEE Transactions on Information Theory}, 52(10):4394--4412, 2006.

\bibitem{NEURIPS2018_288cc0ff}
Zinan Lin, Ashish Khetan, Giulia Fanti, and Sewoong Oh.
\newblock {P}ac{G}{A}{N}: {T}he power of two samples in generative adversarial networks.
\newblock In S.~Bengio, H.~Wallach, H.~Larochelle, K.~Grauman, N.~Cesa-Bianchi, and R.~Garnett, editors, {\em Advances in Neural Information Processing Systems}, volume~31. Curran Associates, Inc., 2018.

\bibitem{mao2017squares}
Xudong Mao, Qing Li, Haoran Xie, Raymond~Y.K. Lau, Zhen Wang, and Stephen Paul~Smolley.
\newblock {L}east squares generative adversarial networks.
\newblock In {\em the IEEE International Conference on Computer Vision (ICCV)}, Oct 2017.

\bibitem{nielsen2020generalization}
Frank Nielsen.
\newblock {O}n a generalization of the {{J}}ensen--{{S}}hannon divergence and the {J}ensen--{S}hannon centroid.
\newblock {\em Entropy}, 22(2):221, 2020.

\bibitem{nielsen2013chi}
Frank Nielsen and Richard Nock.
\newblock {O}n the chi square and higher-order chi distances for approximating f-divergences.
\newblock {\em IEEE Signal Processing Letters}, 21(1):10--13, 2013.

\bibitem{nowozin2016fgan}
Sebastian Nowozin, Botond Cseke, and Ryota Tomioka.
\newblock f-gan: {T}raining generative neural samplers using variational divergence minimization.
\newblock {\em Advances in Neural Information Processing Systems}, 29, 2016.

\bibitem{osterreicher1996class}
Ferdinand {\"O}sterreicher.
\newblock {O}n a class of perimeter-type distances of probability distributions.
\newblock {\em Kybernetika}, 32(4):389--393, 1996.

\bibitem{pan2021exploiting}
Xingang Pan, Xiaohang Zhan, Bo~Dai, Dahua Lin, Chen~Change Loy, and Ping Luo.
\newblock {E}xploiting deep generative prior for versatile image restoration and manipulation.
\newblock {\em IEEE Transactions on Pattern Analysis and Machine Intelligence}, 44(11):7474--7489, 2021.

\bibitem{Radford2015}
Alec Radford, Luke Metz, and Soumith Chintala.
\newblock {U}nsupervised representation learning with deep convolutional generative adversarial networks.
\newblock In {\em Proceedings of the 9th {I}nternational {C}onference on {I}mage and {G}raphics}, pages 97--108, 2017.

\bibitem{renyi1961measures}
Alfr{\'e}d R{\'e}nyi.
\newblock {O}n measures of entropy and information.
\newblock In {\em the Fourth Berkeley Symposium on Mathematical Statistics and Probability, Volume 1: Contributions to the Theory of Statistics}, volume~4, pages 547--562. University of California Press, 1961.

\bibitem{Sason_2018}
Igal Sason.
\newblock {O}n f-divergences: {I}ntegral representations, local behavior, and inequalities.
\newblock {\em Entropy}, 20(5):383, May 2018.

\bibitem{van2014renyi}
Tim Van~Erven and Peter Harremos.
\newblock {R}{\'e}nyi divergence and {K}ullback-{L}eibler divergence.
\newblock {\em IEEE Transactions on Information Theory}, 60(7):3797--3820, 2014.

\bibitem{welfert2023addressing}
Monica Welfert, Gowtham~R. Kurri, Kyle Otstot, and Lalitha Sankar.
\newblock Addressing {G}{A}{N} training instabilities via tunable classification losses, 2023.

\bibitem{sankar2023}
Monica Welfert, Kyle Otstot, Gowtham~R Kurri, and Lalitha Sankar.
\newblock $(\alpha_{D},\alpha_{G})$-{GAN}s: {A}ddressing {GAN} training instabilities via dual objectives.
\newblock In {\em Proceedings of the IEEE International Symposium on Information Theory (ISIT)}, 2023.

\end{thebibliography}

\end{document}